\documentclass[preprint,12pt]{elsarticle}

\usepackage{amssymb,amsmath,amsthm,array,mathtools}
\usepackage{graphicx}
\usepackage{booktabs,multirow}
\usepackage{rotating}
\usepackage{url}

\newtheorem{proposition}{Proposition}[section]
\newtheorem{lemma}{Lemma}
\theoremstyle{remark}
\newtheorem{remark}{Remark}[section]

\newcommand{\h}{\mathbf{h}}
\newcommand{\x}{\mathbf{x}}
\newcommand{\z}{\mathbf{z}}

\renewcommand{\b}{\mathbf{b}}
\newcommand{\W}{\mathbf{W}}

\renewcommand{\H}{\mathbf{H}}
\newcommand{\M}{\mathbf{M}}
\newcommand{\R}{\mathbb{R}}

\journal{Neurocomputing}

\begin{document}

\begin{frontmatter}

\title{Memristive-Friendly Hadamard Reservoir Computing: Structured,
Multiplier-Free Recurrences at Scale%
}

\author[pisa]{Andrea Ceni}
\author[inrim]{Gianluca Milano}
\author[polito]{Carlo Ricciardi}
\author[pisa]{Claudio Gallicchio\corref{cor1}}
\ead{claudio.gallicchio@unipi.it}
\cortext[cor1]{Corresponding author}

\affiliation[pisa]{organization={Department of Computer Science, University of Pisa}, country={Italy}}
\affiliation[inrim]{organization={Istituto Nazionale di Ricerca Metrologica (INRiM)}, country={Italy}}
\affiliation[polito]{organization={Department of Applied Science and Technology, Politecnico di Torino}, country={Italy}}

\begin{abstract}
Reservoir Computing (RC) designs Recurrent Neural Networks around a
fixed, i.e., untrained, recurrent layer, and is a natural candidate for
neuromorphic hardware. Memristive-friendly reservoirs derive the
neuron dynamics from memristive-device kinetics, but still rely on dense
recurrent matrices, which are expensive to realize physically. In this paper, we
replace the dense matrix with a structured orthogonal operator, built
from sign diagonals, a permutation, and a fast Walsh-Hadamard transform.
The operator is multiplier-free, requires $O(N)$ parameters and
$O(N\log N)$ operations per step, and is never materialized as a matrix.
We instantiate it in a standard and in a memristive-friendly Echo State
Network, with one binary input connection per unit.

Our mathematical analysis shows that exact orthogonality yields an echo
state condition that is tight in the recurrent scaling, and a noise
response that is predictable at design time. Moreover, the operator mixes
the whole state in a single application.
Experiments on twenty classification and seven regression benchmarks, at
reservoir sizes up to $N = 8192$, show that the structured models match dense
orthogonal reservoirs, and 
achieve better mean performance than the cycle reservoir
by a
margin that widens with size. Furthermore, we time the recurrent step
on three hardware platforms, where it is up to $50\times$ faster than a
dense product and $10^4\times$ smaller in memory. Finally, we ablate the
operator and measure the response to noise, quantization, device mismatch
and discrete faults.
\end{abstract}

\begin{keyword}
Reservoir Computing 
\sep Echo State Networks 
\sep Hadamard transform 
\sep Neuromorphic computing 
\end{keyword}

\end{frontmatter}

\section{Introduction}
\label{sec:intro}
Neuromorphic computing seeks to overcome the energy and bandwidth
limitations of conventional digital architectures by co-locating memory
and processing in physical substrates whose intrinsic dynamics carry out
the computation \cite{sangwan2020neuromorphic,sebastian2020memory}.
Memristive devices provide a fundamental pillar of this program: their conductance
is modulated by the history of the applied stimuli, providing a compact
analog state variable with synapse-like plasticity
\cite{miranda2020modeling,milano2022memristive}, and self-organized
networks of such devices have been shown to process temporal information
\emph{in materia}, directly in the physics of the substrate
\cite{milano2022materia,milano2023nanowire}.
On the algorithmic side, the natural partner of these substrates is
Reservoir Computing (RC) \cite{lukovsevivcius2009reservoir,nakajima2021reservoir,yan2024emerging}, the paradigm in which a recurrent neural layer, the \emph{reservoir}, is left untrained and acts as a fixed nonlinear filter of the input history,
while learning is confined to a (typically linear) readout. 
Precisely because the recurrent part need not be adapted, it can be delegated to a physical dynamical system, and RC has become the reference framework for computing with physical dynamics \cite{tanaka2019recent}. Bridging the two levels, the recently proposed memristive-friendly Echo State Network (MF-ESN)
\cite{pistolesi2025memristive} derives the neuron update of an otherwise
standard reservoir from the potentiation-depression kinetics of real
memristive devices, bringing the reservoir abstraction one step closer to
a physical implementation.

Neuron-level fidelity, however, does not remove the main structural
obstacle to that implementation, namely the recurrent \emph{coupling}. 
Indeed, in the Echo State Network (ESN) architecture, the reservoir units are
often coupled by means of a dense random matrix, which entails $O(N^2)$ weights to store, $O(N^2)$ multiply-accumulate operations per step and, in a
physical realization, $O(N^2)$ distinct programmable conductances with the associated precision and calibration requirements. 
This impact is heaviest precisely in the places where RC is most appealing: predictive quality grows with the reservoir size $N$, so the regime of practical interest is the one in which a dense coupling is least affordable.
The hardware-oriented alternatives explored so far sit at two extremes
(Section~\ref{sec:related}). On one side we find minimal deterministic
topologies, such as the ring of the Simple Cycle Reservoir (SCR)
\cite{rodan2011minimum}, which maps directly onto a circular shift register (a delay loop), but through which information travels one position per step. On the other we find dense \emph{orthogonal} recurrences, which maximize short-term memory \cite{white2004short,farkas2016memory}, but are as expensive to store and to apply as random ones. Between the two lies the family of structured orthogonal operators built from fast
transforms, which approximate dense random projections at $O(N\log N)$
multiplier-free cost \cite{le2013fastfood,yu2016orthogonal} and have
entered RC as efficient \emph{input} projections and random feature maps
\cite{dong2020reservoir}. However, the systematic use of this family as
the \emph{recurrent} operator of a reservoir is still missing. In that
role, its exact orthogonality affects stability and memory, its global
mixing affects the quality of the state representation, and its
multiplier-free structure affects physical realizability.

In this paper, we fill this gap by introducing and analyzing
\emph{Hadamard reservoirs}, i.e., recurrent
networks whose coupling is the structured orthogonal operator
$\M = \mathbf{D}_2\; \widehat\H_N\; \Pi\; \mathbf{D}_1$, built from two
diagonal sign matrices, a permutation, and a normalized Walsh-Hadamard
transform. The operator is applied in $O(N\log N)$ additions and
subtractions, stores $O(N)$ parameters, and is never materialized as a
matrix. We develop the construction along two tracks. First, we use the
operator as a general hardware-friendly primitive, replacing the dense
recurrence of a standard leaky ESN \cite{jaeger2007optimization}, which
yields the H-ESN. Then, we bring it to the neuromorphic setting,
replacing the coupling of the memristive-friendly model, which yields
the MF-H-ESN, whose neuron dynamics remain those of memristive devices
while the recurrence reduces to sign flips, fixed routing, and pairwise
additions.
In both cases the substitution concerns the coupling alone
and leaves the neuron model untouched.
A further simplification of the input pathway, i.e., a single binary
input connection per unit, completes a model family in which every
analog-valued weight matrix has been removed from the architecture.
The reason why so much structure can be imposed at so little functional
cost is representational, and runs through the whole paper. The Hadamard
operator inherits the uniform memory of dense orthogonal recurrences
\emph{and} the one-step global mixing of dense random ones. A coordinate
impulse spreads over the entire state in a single application, at the
hardware complexity of a structured, multiplier-free operator. Moreover,
its exact orthogonality has direct consequences for the analysis: the echo state condition becomes
tight in the single tuned scalar, and the response to runtime
perturbations becomes predictable at design time.

The contributions of this work can be summarized as
follows.\footnote{This paper is an extended version of
\cite{ceni2026hadamard}.}
\begin{itemize}
\item We introduce the H-ESN and the MF-H-ESN, whose recurrent coupling is
the multiplier-free structured orthogonal operator $\M$, together with
input pathways reduced to one binary connection per unit
(Section~\ref{sec:model}).

\item We derive their echo state condition, which is \emph{tight} in the
recurrent scaling $\rho_M$, and a noise gain that is fixed at design time;
a spectral and mixing analysis isolates what each randomization element
contributes (Section~\ref{sec:theory}).

\item We measure what the recurrent step costs, in wall-clock time on
three hardware platforms and in memory, and find that the crossover with
the dense product is a property of the device, while the memory advantage
is not (Section~\ref{sec:implementation}).

\item We evaluate the models on twenty classification and seven regression
benchmarks, against dense, cycle, dense-orthogonal and fully trainable
gated recurrent baselines, up to $N = 8192$, with a critical-difference
analysis (Section~\ref{sec:experiments}).

\item We ablate the operator, and show that the permutation protects
memory near the edge of stability, and the sign diagonals the
representational richness of the nonlinear states
(Section~\ref{sec:ablation}).

\item We assess robustness to state noise, in-loop quantization, device
mismatch, and faults in the signs and in the routing, and we report what
each of them costs in accuracy (Section~\ref{sec:robustness}).
\end{itemize}

The rest of this paper is organized as follows.
Section~\ref{sec:background} recalls ESNs, the memristive-friendly
model, and related work on structured and hardware-friendly reservoir
topologies. Section~\ref{sec:model} introduces the Hadamard reservoir
models and analyzes their computational and hardware cost.
Section~\ref{sec:implementation} then measures what the recurrent step
costs, in wall-clock time across hardware platforms and in memory.
Section~\ref{sec:theory} develops the theoretical analysis, and
Section~\ref{sec:experiments} reports the empirical evaluation.
Section~\ref{sec:discussion} then puts the proposed models side by side
with the reservoir models taken as reference. Finally,
Section~\ref{sec:conclusions} concludes the paper.

\section{Background and related work}
\label{sec:background}

In this section we provide the background of the paper.
First, in Section~\ref{sec:esn} we recall the ESN formulation.
Then, in Section~\ref{sec:mfesn} we describe the memristive-friendly
reservoir model from which the neuron dynamics of this work are derived.
Finally, in Section~\ref{sec:related} we place the structured recurrence
we propose within the existing landscape of hardware-oriented reservoir
topologies.

\subsection{Echo State Networks}
\label{sec:esn}
Echo State Networks (ESNs)~\cite{jaeger2004harnessing} 
provide a foundational and popular instance of the RC concept: an untrained layer of recurrent non-linear neurons coupled with a trainable layer of linear readout neurons.
In the following, we adopt the leaky ESN formulation from~\cite{jaeger2007optimization} to develop the mathematical description.
Let $\h_t\in \R^N$ and $\x_t\in\R^d$ denote, respectively, the reservoir
state and the external input at time $t$. Reservoir dynamics are described in terms of an input-driven non-linear iterated map, as follows:
\begin{equation}
\label{eq.esn_reservoir}
\h_t = (1-\alpha) \h_{t-1} + \alpha \tanh{\big( \W_h \h_{t-1} + \W_x \x_t + \b_h \big)},
\end{equation}
where $\W_h\in\R^{N\times N}$ and $\W_x\in\R^{N\times d}$ are,
respectively, the recurrent weight matrix and the input weight matrix,
while $\b_h\in\R^N$ is a bias vector. Moreover, $\tanh(\cdot)$ indicates
an element-wise applied hyperbolic tangent non-linearity. Finally,
$\alpha\in (0,1]$ is the leaking rate, which influences the speed of the
reservoir update with respect to the dynamics in the input signal.
All the weights in $\W_h$, $\W_x$, and $\b_h$ are left untrained after initialization under stability constraints given by the Echo State Property (ESP) \cite{jaeger2004harnessing,yildiz2012re}. This essentially ensures that the network's state is determined by the input history rather than by initial conditions. 
A commonly adopted strategy for reservoir initialization is to draw the recurrent weights in $\W_h$ randomly from a uniform distribution in $(-1,1)$ and then re-scale them to meet a desired value $\rho$ of the spectral radius (i.e., the maximum among the absolute values of the eigenvalues) of $\W_h$.
Values of $\rho < 1$ are typically explored in accordance with a necessary
condition for the ESP (see \cite{jaeger2001echo} for details), although
this condition is not sufficient in general, and guaranteeing the property
for every input requires the more restrictive constraint
$\|\W_h\|_2 < 1$, which describes a conservative regime rarely adopted in
practice \cite{jaeger2001echo,yildiz2012re}. The value of $\rho$ is
therefore treated as a hyper-parameter, with settings closer to unity
corresponding to a longer memory of the input history, and smaller values
to a faster forgetting. The entries of the input matrix $\W_x$ and of the
bias $\b_h$ are drawn from uniform distributions in
$(-\omega_x, \omega_x)$ and $(-\omega_b, \omega_b)$. The two scaling
factors $\omega_x$ and $\omega_b$, together
with $\rho$ and the leaking rate $\alpha$, form the set of hyper-parameters that shape the
reservoir dynamics.

\subsection{Memristive-friendly Reservoir Computing}
\label{sec:mfesn}
A memristive-friendly formulation of the ESN operation has been introduced
in \cite{pistolesi2025memristive}, where the state update is derived from the effective kinetics of memristive devices, through a
potentiation-depression mechanism governed by rates
$K_p(z) = \kappa_{p0} e^{\eta_p z}$ and $K_d(z) = \kappa_{d0}e^{-\eta_d z}$:
\begin{align}
\z_{t} &= \text{Rescale}(\W_h \h_{t-1} + \W_x \x_t + \b_h), \label{eq.mf_linear}\\
\h_{t} &= \gamma \h_{t-1} + \varepsilon \big[ K_p(\z_{t})
- \text{diag}( K_p(\z_{t}) + K_d(\z_{t}) ) \mathbf{h}_{t-1}
\big],\label{eq.mf_reservoir_h}
\end{align}
where $\text{Rescale}(z) = (b - a)/(1 + e^{-z s}) + a$, with $a = 0.35$,
$b = 1.15$ fixed as in \cite{pistolesi2025memristive}, and $s$ a
hyper-parameter tuning the nonlinearity of the rescaling.

The two equations have a physical reading. The Rescale function
maps the linear pre-activation into the voltage window $[a, b]$ in which
the device is operated, with the hyper-parameter $s$ controlling the
steepness of the mapping, hence how sharply the drive is compressed
towards the two ends of the window. The rescaled value $z_{t,i}$ plays
the role of the voltage applied to the $i$-th device, and the state
$h_{t,i}$ that of its normalized conductance. The rate $K_p$ grows with
the applied voltage and drives potentiation, while $K_d$ decreases with
it and drives depression, with $\kappa_{p0}$ and $\kappa_{d0}$ setting
the two base rates and $\eta_p$ and $\eta_d$ their sensitivity to the
voltage. With the constants typical of memristive nanowire networks
\cite{milano2022connectome}, potentiation is negligible at the lower end
of the window and dominant at the upper end, so that the effective
non-linearity of the neuron arises from the competition between the two
rates rather than from an explicitly chosen activation function.

In eq.~\ref{eq.mf_reservoir_h}, $\varepsilon$ acts as the integration
step of the underlying kinetics, i.e., as the amount of conductance change
produced by a single input presentation. The coefficient
$\gamma \in (0,1]$ accounts instead for the volatility of the device,
with $\gamma = 1$ describing an ideal non-volatile behavior, and smaller
values a state that decays on its own. Past information is thus carried forward with the effective retention
factor $\gamma - \varepsilon S(\z_t)$, where $S = K_p + K_d$, so that
$\gamma$ and $\varepsilon$ play jointly the role that the leaking rate
$\alpha$ plays in eq.~\ref{eq.esn_reservoir}, and they control the
contraction of the state map (see details in Section~\ref{sec:theory}). Among the
quantities above, $\varepsilon$, $\gamma$ and $s$ are treated as
hyper-parameters, while $a$, $b$ and the rate constants are fixed 
(see details in~\ref{app:protocol}).

\subsection{Structured and hardware-friendly reservoir topologies}
\label{sec:related}

The observation that effective reservoir computation does not require
unconstrained random connectivity has a long history. Rodan and Ti\v{n}o \cite{rodan2011minimum} showed that reservoirs with minimal, fully deterministic topology perform comparably to standard ESNs, while admitting an exact characterization of their memory capacity. 
The most prominent instance is the SCR \cite{rodan2011minimum}, a ring of identical weights, with input connections of a single magnitude whose signs are fixed deterministically by an aperiodic sequence.
Ring-like
topologies have since become a de facto standard in physical reservoir
computing, as a cycle requires minimal routing resources and is naturally
realized by a single node with delayed feedback \cite{tanaka2019recent}. At the opposite end of the spectrum, dense \emph{orthogonal} recurrent
matrices are known to maximize short-term memory capacity
\cite{white2004short,farkas2016memory}, and orthogonal or unitary
constraints have been exploited in fully trained recurrent networks as
well, to keep gradients from vanishing over long horizons
\cite{arjovsky2016unitary,ceni2025random}. A generic orthogonal matrix, however, is as
expensive to store and to apply as a dense random one, it has to be
generated by an $O(N^3)$ factorization, e.g., the QR decomposition of a
random matrix, and it offers no route to physical simplification.

The family of structured orthogonal operators built from fast
transforms lies between these two extremes, i.e., between maximally
local structure and maximally unstructured orthogonality. Fastfood \cite{le2013fastfood} and
structured orthogonal random features \cite{yu2016orthogonal} established
that products of diagonal sign matrices, permutations, and Walsh-Hadamard
transforms approximate dense Gaussian and orthogonal projections at
$O(N \log N)$ cost, and structured transforms have been brought to
RC in the context of optical random features and
recurrent kernels \cite{dong2020reservoir}. Optimized FWHT kernels have
since been written outside RC, for the Hadamard-based quantization of large
language models \cite{dao2023fht,hadacore2024,ashkboos2024quarot}, and the
recurrence studied here can use them directly. The operator adopted in this
work belongs to this family; differently from \cite{dong2020reservoir},
we employ it as the \emph{recurrent} operator of an otherwise standard, or
memristive-friendly, ESN, and we focus on its consequences for hardware
realizability, stability, and scaling behavior.

Finally, the cycle topology also admits a sharp spectral characterization:
linear SCRs at the edge of stability have recently been shown to represent
the driving signal in the Fourier basis \cite{tino2024linear}, a result
that rests on the exact periodicity of the cyclic shift. The operator
proposed here trades that periodicity for global mixing, whose
consequences we analyze in Section~\ref{sec:theory}.

\section{Hadamard reservoirs}
\label{sec:model}

In this section we introduce the proposed models.
First, in Section~\ref{sec:operator_def} we define the structured
orthogonal operator that replaces the dense recurrent matrix.
Then, in Section~\ref{sec:model_defs} we build two reservoir models on
it, i.e., the H-ESN and the memristive-friendly MF-H-ESN, together with
the simplified input pathway that completes the construction.
Finally, Sections~\ref{sec:cost} and~\ref{sec:hardware} quantify the
resulting computational cost and discuss what it implies for a physical
implementation.

\subsection{Structured orthogonal recurrence}
\label{sec:operator_def}
For a reservoir state dimension $N=2^K$, the Hadamard matrix
$\H_N\in\{\pm1\}^{N\times N}$ is defined recursively by
$\H_1 = [1]$, $\H_{2n} = \begin{bsmallmatrix} \H_n & \H_n \\ \H_n & -\H_n \end{bsmallmatrix}$,
and the normalized transform $\widehat \H_N = \H_N/\sqrt{N}$ is orthogonal.
We replace the dense recurrent matrix with the structured orthogonal operator defined as follows:
\begin{equation}
\label{eq.operator}
\M = \mathbf{D}_2 \; \widehat \H_N \; \Pi \; \mathbf{D}_1,
\end{equation}
where $\mathbf{D}_1, \mathbf{D}_2$ are diagonal matrices with random
independent $\pm 1$ entries and $\Pi$ is a random permutation matrix.
Being a product of orthogonal factors, $\M$ is exactly orthogonal, i.e.,
$\|\M \h\| = \|\h\|$ for every $\h$. The three randomization elements
inject controlled randomness and heterogeneity while preserving this
property, and each of them is necessary, for reasons that we analyze in
Section~\ref{sec:theory} and isolate empirically in the ablation of
Section~\ref{sec:ablation}.
In the models of Section~\ref{sec:model_defs} the recurrent coupling is
the scaled operator $\rho_M \M$, where the single scalar $\rho_M > 0$
controls the strength of the recurrence.

\begin{figure}[!htbp]
\centering
\includegraphics[width=\linewidth]{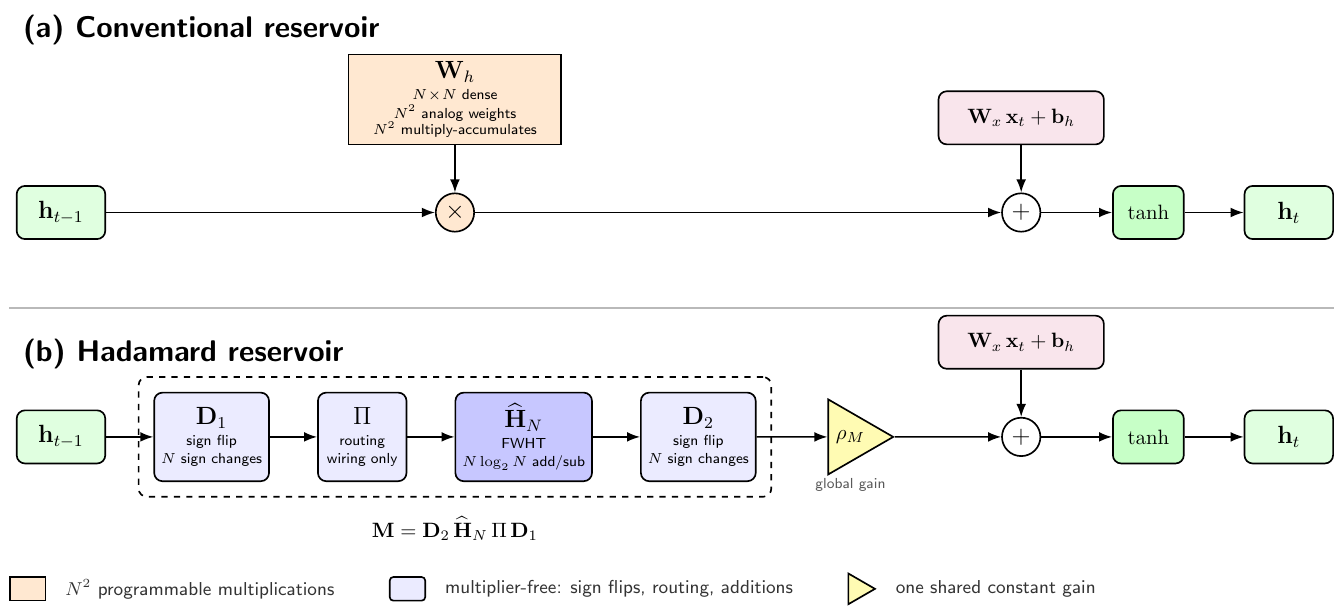}
\caption{Comparison between the recurrent step of a conventional
reservoir (a) and of the Hadamard reservoir proposed in this work (b).
In (a) the dense matrix $\W_h$ stores $N^2$ analog weights and its
application costs $N^2$ multiply-accumulate operations. In (b) the same
state mixing is obtained from the structured operator
$\M = \mathbf{D}_2 \widehat\H_N \Pi \mathbf{D}_1$ (eq.~\ref{eq.operator}),
applied as a sequence of sign flips ($\mathbf{D}_1$, $\mathbf{D}_2$), a
fixed routing ($\Pi$), and a fast Walsh-Hadamard transform ($\widehat\H_N$,
detailed in \ref{app:implementation}). The stored description reduces to the
$2N$ signs of the two diagonals, the $N$ indices of the permutation and
the single scalar $\rho_M$, and the $N\log_2 N$ operations are additions
and subtractions. No programmable multiplication is involved, as the
recurrence is scaled by one constant gain, common to all the units, which
can be absorbed into the entries of $\mathbf{D}_2$.}
\label{fig:concept}
\end{figure}

Figure~\ref{fig:concept} contrasts the resulting recurrent step with a
conventional dense reservoir. Replacing the dense matrix-vector product
$\W_h \h(t-1)$ with the four-stage pipeline $\mathbf{D}_1 \to \Pi \to
\widehat\H_N \to \mathbf{D}_2$ turns an $O(N^2)$ full-precision analog
multiplication into a sequence of sign flips, a fixed routing, and a
multiplier-free transform. The recurrent weights reduce to the
scalar $\rho_M$, together with the random signs and the permutation drawn
at initialization.

\subsection{Model definitions}
\label{sec:model_defs}
The \textbf{H-ESN} is obtained from eq.~\ref{eq.esn_reservoir} by
replacing the dense recurrence $\W_h$ with the scaled structured operator:
\begin{equation}
\label{eq.hesn}
\h_t = (1-\alpha)\, \h_{t-1} + \alpha \tanh\big( \rho_M \, \M\, \h_{t-1}
+ \W_x \, \x_t + \b_h \big).
\end{equation}
The \textbf{MF-H-ESN} applies the same substitution to the linear
pre-activation of the memristive-friendly update, eq.~\ref{eq.mf_linear}:
\begin{equation}
\label{eq.mfhesn}
\z_{t} = \text{Rescale}\big(\rho_M \,\M \,\h_{t-1} + \W_x \, \x_t + \b_h\big),
\end{equation}
with the state update unchanged from eq.~\ref{eq.mf_reservoir_h}.
In both models the single scalar $\rho_M$ controls the strength of the
recurrence and plays the role of the spectral radius. It does so exactly: indeed, every eigenvalue of $\rho_M \M$ has modulus $\rho_M$
(see details in Section~\ref{sec:theory}).

\begin{figure}[!htbp]
\centering
\includegraphics[width=\linewidth]{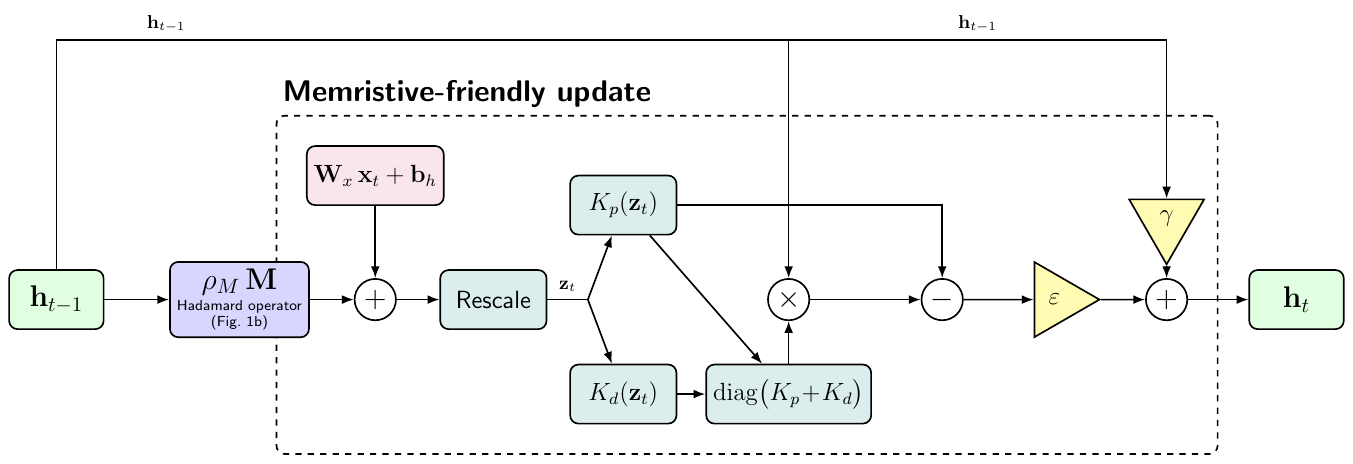}
\caption{Computational graph of the memristive-friendly Hadamard reservoir
(MF-H-ESN): the linear pre-activation of eq.~\ref{eq.mf_linear} is replaced
by the Hadamard version of eq.~\ref{eq.mfhesn}, while the state update
eq.~\ref{eq.mf_reservoir_h} is unchanged. The dense recurrent term of the
original memristive-friendly update \cite{pistolesi2025memristive} is
replaced by the structured Hadamard operator $\rho_M\M$ (compact box,
detailed in Figure~\ref{fig:concept}b), which feeds the
potentiation/depression kinetics $K_p, K_d$ and the leaky integration
(dashed box) exactly as in the original model.}
\label{fig:mfhadamard}
\end{figure}

Figure~\ref{fig:mfhadamard} depicts the resulting MF-H-ESN state update: the
dense recurrent term of the original memristive-friendly update
\cite{pistolesi2025memristive} is replaced by the compact Hadamard pipeline
of Figure~\ref{fig:concept}b, while the potentiation-depression kinetics
$K_p, K_d$ that give the model its memristive character are otherwise
untouched. The structured operator only affects \emph{how} the previous
state is mixed before entering the memristive nonlinearity, not the
nonlinearity itself.

The input pathway of both models is initialized as in the conventional
leaky ESN of Section~\ref{sec:esn}, with the entries of $\W_x$ and
$\b_h$ drawn uniformly in $(-\omega_x, \omega_x)$ and
$(-\omega_b, \omega_b)$. 
We further consider a hardware-oriented simplified-input variant, denoted by the subscript $_\mathrm{si}$ (for ``simplified input''), in which each reservoir
unit is wired to a single input channel, chosen
uniformly at random, with weight $\pm\omega_x$, and the bias entries are
$\pm\omega_b$. This removes every analog-valued parameter from the input
path and reduces the input wiring from $N d$ to $N$ physical connections,
one per unit.

\subsection{Cost analysis}
\label{sec:cost}
Interestingly, the structured operator $\M$ is applied without ever storing the full
matrix: the diagonal matrices introduce sign changes, the permutation
reorders coordinates, and the Hadamard transform is computed by the fast
Walsh-Hadamard transform (FWHT) in $O(N\log N)$ additions and
subtractions, via the radix-2 butterfly decomposition of $\widehat\H_N$
(Figure~\ref{fig:butterfly} in \ref{app:implementation}). This
reduces the cost of the recurrent step from $O(N^2)$ to $O(N \log N)$
operations, and the storage from $O(N^2)$ to $O(N)$ parameters, i.e., $2N$
signs, $N$ routing indices and a scaling constant. Those parameters are
discrete, and a routing index takes $\log_2 N$ bits, so that the operator
is described by $O(N\log N)$ bits in total, against the $O(N^2)$ real
values of a dense matrix; Section~\ref{sec:memory} turns both counts into
explicit figures. Moreover, the operator is
\emph{multiplier-free}: apart from the global scalings, it involves only
additions, subtractions, sign flips, and routing.
A corresponding reduction applies to the simplified input pathway described in
Section~\ref{sec:model_defs}: $\W_x$ and $\b_h$ no longer require
storing floating-point entries, as they are defined by deterministic
scalings of binary patterns, and the input wiring collapses to one
connection per unit, with a single global gain.

\subsection{Hardware considerations}
\label{sec:hardware}
The decomposition of the recurrence into sign flips, routing, and
pairwise additive interactions substantially lowers the complexity of a
physical implementation compared to a dense analog matrix. In contrast to
a conventional dense recurrent weight matrix, which requires $O(N^2)$
distinct conductances and analog multiplications, the operator $\M$ can
be realized by structured interconnects and polarity-controlled
summations, simplifying the memristive hardware architecture  \cite{sebastian2020memory}. The simplified input structures further reduce the number of analog
parameters. In the fully simplified models, both the recurrent and the
input operations rely entirely on polarity manipulations and simple
scalings, which align naturally with the constraints of memristive
hardware, and retain the global mixing ensured by the structured
recurrence.
In this work we study the algorithmic side of this design, i.e., the
dynamical, computational and robustness consequences of the structured
recurrence. A circuit-level design, and a physical realization of the
butterfly operator on a memristive substrate, are left to future work.

\section{Computational cost and memory footprint}
\label{sec:implementation}

This section quantifies the two costs of the recurrent step. The first
one, wall-clock time, is a property of the operator and of the machine
that runs it, and we measure it on three hardware platforms. The second
one, memory, does not depend on the machine, and it is the quantity that
decides whether a reservoir of a given size can be physically
instantiated at all.

\subsection{Wall-clock time across hardware platforms}
\label{sec:timing}

The cost analysis of Section~\ref{sec:cost} shows that the structured
operator computes the recurrent step in $O(N \log N)$ additions and
subtractions, against $O(N^2)$ multiply-accumulates for a dense one. That
count is asymptotic, and it leaves open the question that matters in
practice, i.e., at which reservoir size the structured step actually
becomes faster than the dense product, and whether that size is a
property of the operator or of the machine it runs on. We answer it
empirically, by timing the recurrent step $\h \mapsto \rho_M \M \h$
against a dense matmul baseline on three platforms that span two orders
of magnitude in raw throughput: a laptop-class Apple~M3~Pro CPU, an
entry-level NVIDIA T4 GPU, and a datacenter-class NVIDIA A100 GPU.

We time $N = 2^8, \ldots, 2^{15}$ in single precision, applying the
recurrence to a batch of $256$ states at once, as happens when $256$
independent sequences are advanced by one time step in parallel, so that
the dense baseline is a $256 \times N$ by $N \times N$ matrix product. We
report wall-clock means over repeated executions after warm-up. The
structured step is realized in three ways, i.e., a plain implementation,
a compiled one, and a dedicated GPU kernel.
Implementation details, and the timings in tabular form, are reported in
\ref{app:implementation}.

\begin{figure}[!htbp]
\centering
\includegraphics[width=\linewidth]{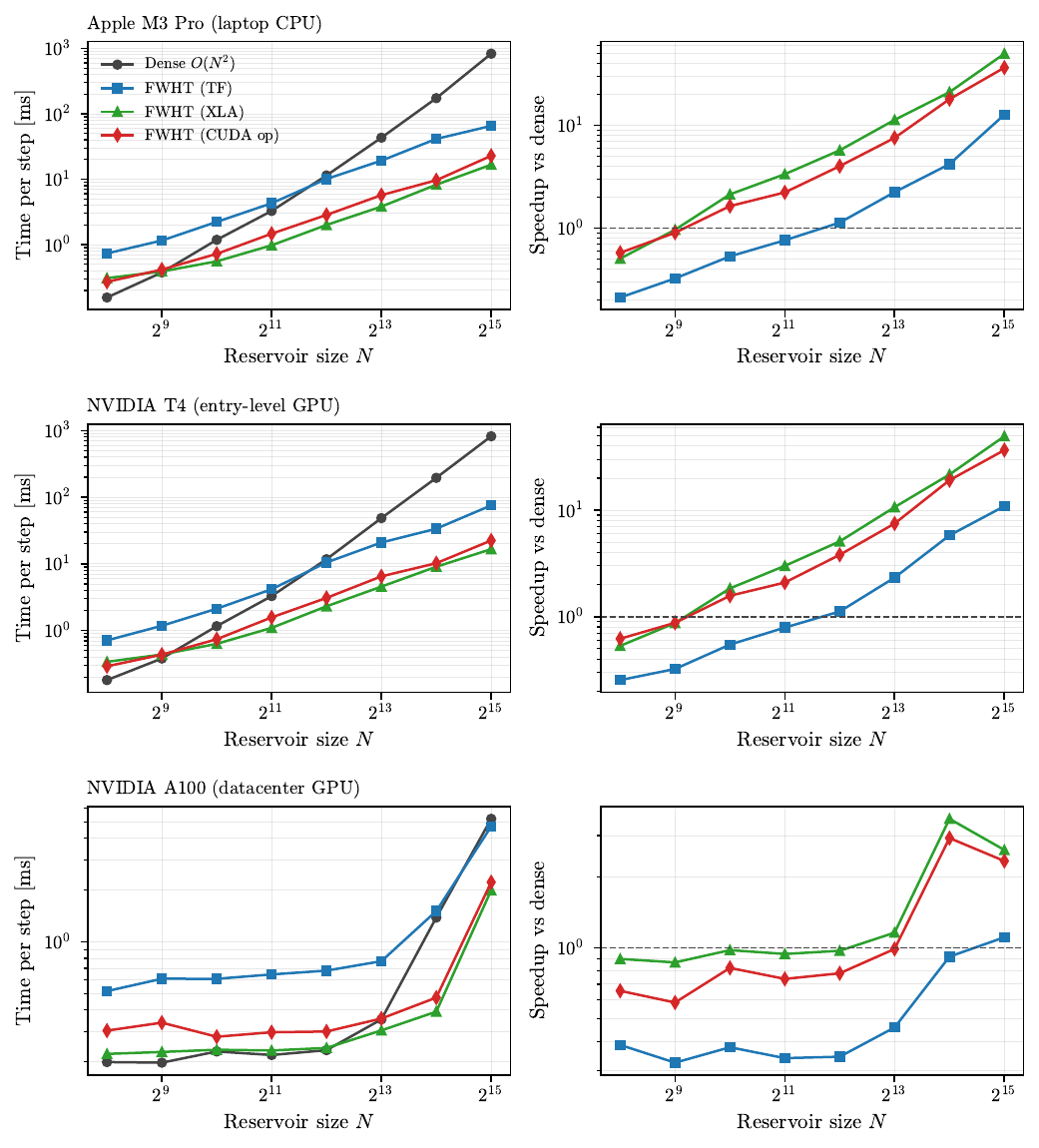}
\caption{Wall-clock time per recurrent step (left) and speedup over the
dense matmul baseline (right) as a function of the reservoir size $N$
(fp32, $256$ states advanced per call), on three hardware platforms. Top: laptop-class
Apple~M3~Pro CPU; middle: entry-level NVIDIA T4 GPU. On both, the
structured recurrence crosses over the dense one at $N\!\approx\!1024$
with a compiled implementation, and reaches about $50\times$ at
$N=32768$. Bottom: datacenter-class NVIDIA A100 GPU, where the crossover
moves to $N\!\approx\!8192$. Below that size all implementations are
dominated by a fixed per-call overhead, and the structured step stays
within about ten percent of the dense one; beyond it the advantage grows
($3.5\times$ at $N=16384$). Timings in tabular form are in
\ref{app:implementation}.}
\label{fig:crossover}
\end{figure}

Figure~\ref{fig:crossover} reports the results, from which three main
observations can be drawn.
First, on the laptop-class CPU and on the T4 the structured step
overtakes the dense product already at $N \approx 1024$, reaching almost
identical peak speedups of about $50\times$ at $N = 32768$. The settings closest to edge deployment, where the
memristive scenario is targeted, are thus precisely those where the
structure pays off earliest.
Second, the crossover point is a property of the device, and scales with
its raw dense-matmul throughput. On the A100, whose tensor cores make
dense products extremely fast, the crossover moves to $N \approx 8192$,
and below that size the structured step stays within about ten percent of
the dense one. The advantage then grows to $3.5\times$ at $N = 16384$.
Third, the operation count alone does not predict the wall-clock time.
The plain implementation performs asymptotically fewer operations than
the dense product, and yet it does not overtake it on the A100 in the
range we tested.

\subsection{Memory footprint}
\label{sec:memory}

\begin{figure}[!htbp]
\centering
\includegraphics[width=\linewidth]{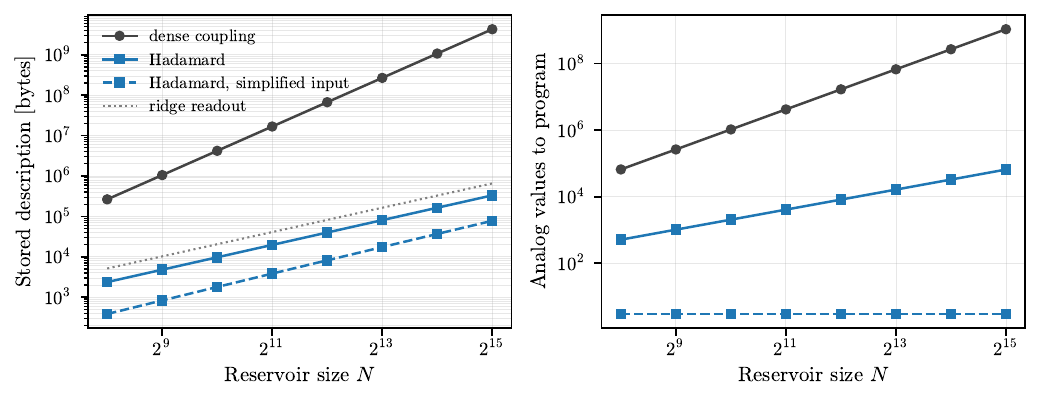}
\caption{Memory footprint of the structured recurrence against the dense
one, on a univariate task ($d = 1$). Left: bits needed to store the
recurrent operator and the input pathway, with real-valued parameters at
$32$ bits, signs at one bit and routing indices at $\log_2 N$ bits. Right:
analog values that a physical realization has to program individually. The
dense curve is the ESN, the dense orthogonal reservoir and the MF-ESN,
which coincide, and each Hadamard curve is a $\tanh$ and a
memristive-friendly model, i.e., H-ESN and MF-H-ESN, and their
simplified-input variants, which coincide in the same way, since they
share the coupling and the input pathway; the analog values of the
simplified-input variants are three at every size.
The dotted line is the ridge readout of a five-class task, which every
reservoir model carries in addition. Formulas, assumptions and the
corresponding table are in \ref{app:params}.}
\label{fig:footprint}
\end{figure}

The size of the timing advantage depends on the platform, as we have just
seen. The memory cost does not, i.e., the description a model has to
store, and the number of components a physical realization has to
instantiate, follow from the operator alone. This is the side of the cost
that decides whether a reservoir of a given width can be built at all, and
it weighs most in the memristive scenario, where the recurrent coupling
occupies device area rather than memory.
Figure~\ref{fig:footprint} compares the structured recurrence with the
dense one, as the previous subsection did for time, on the two counts that
matter, i.e., the description to be stored and the analog values that have
to be programmed one by one. The formulas behind the two panels, the
assumptions of the count, and the corresponding table are reported in
\ref{app:params}.

A dense coupling stores $32N^2$ bits and requires as many independently
programmed analog values, i.e., $268$~MB and $6.7 \times 10^7$ values at
$N = 8192$. This is the price of unstructured mixing, and it is what puts
the largest sizes of Section~\ref{sec:scaling} out of reach of a physical
substrate. The structured operator stores instead $2N$ signs, $N$ routing
indices and one gain, i.e., $15.4$~kB at the same size, a reduction by a
factor $32N/(2 + \log_2 N)$ that grows essentially linearly with the
reservoir size and reaches $1.7 \times 10^4$ there. Put differently, the
memory that a dense recurrence occupies at $N = 256$, the reference size of
this study, would hold a structured one of more than $10^5$ units. Both
counts leave out the handful of global hyper-parameters that every model
carries, e.g., the leaking rate, which do not grow with $N$
(\ref{app:params}).

The right panel carries the point that matters most for a physical
realization. The analog values to be programmed drop from $N^2$ to one,
i.e., the gain $\rho_M$, which can itself be absorbed into $\mathbf{D}_2$
or realized as a single amplifier common to all the units, and with the
simplified input pathway they drop to three global scalars, independent of
the reservoir size. Everything else is discrete, i.e., signs, routing
indices, and the unit-gain connections of the butterfly stages. The dotted
line is the ridge readout, which is common to all the reservoir models and
is not our concern here, beyond one remark: once the recurrence is
structured, the readout becomes the largest object in the model, so that
the memory budget is set by the task rather than by the reservoir
(\ref{app:params}).

Two points carry over to the rest of the paper. First, the accuracy
reported in Section~\ref{sec:experiments} is obtained at a cost that is
lower by four orders of magnitude, so the structure is not a concession
made for the sake of hardware. Second, a description made of $2N$ signs,
$N$ routing indices and one gain is small and almost entirely discrete,
which is what makes the device difficult to corrupt, as
Section~\ref{sec:robustness} quantifies.

\section{Theoretical analysis}
\label{sec:theory}

Having established what the structured recurrence costs, we now turn to
what it does. This section proceeds in three steps. First, in
Section~\ref{sec:theory_esp} we analyze the stability of the MF-H-ESN
dynamics, and derive a \emph{sufficient} condition for the
ESP that is \emph{tight} in the recurrent scaling
$\rho_M$: thanks to the exact orthogonality of the structured operator,
there is no gap between the quantity that is controlled at design time
and the quantity that governs contraction. Then, in
Section~\ref{sec:theory_mixing} we move to the quality of the state
representation, contrasting the spectral and mixing properties of the
Hadamard operator with those of dense reservoirs and of the SCR, and
isolating the role of the randomization
elements $\mathbf{D}_1, \Pi, \mathbf{D}_2$; each claim is accompanied
there by the measurement that supports it, on synthetic signals at
$N = 256$. Finally, in Section~\ref{sec:theory_diagnostics} we verify the
noise response of the recurrence numerically.

\subsection{Echo state property condition}
\label{sec:theory_esp}
We study the MF-H-ESN reservoir as a discrete-time dynamical system,
whose state update we write as $\h_t = \Phi(\h_{t-1}, \x_t)$, with
\begin{equation}
\label{eq.phi}
\Phi(\h, \x) = \big(\gamma \mathbf{1} - \varepsilon S(\z)\big) \odot \h
+ \varepsilon K_p(\z),
\qquad
\z = R\big(\rho_M \M \h + \W_x \x + \b_h\big),
\end{equation}
where $S = K_p + K_d$ and $R$ denotes the Rescale function, which maps
$\R$ into $(a,b)$ and is Lipschitz with constant $L_R = s(b-a)/4$.

Our analysis rests on a single fact, i.e., that $\z$ takes values in
$(a,b)$ componentwise. From it, a few elementary bounds follow. On this interval the rates
are uniformly bounded, $K_p(\z) \in [K_p^-, K_p^+]$ and
$K_d(\z) \in [K_d^-, K_d^+]$, with extremes attained at the endpoints:
$K_p^+ = \kappa_{p0}e^{\eta_p b}$, $K_p^- = \kappa_{p0}e^{\eta_p a}$,
$K_d^+ = \kappa_{d0}e^{-\eta_d a}$, $K_d^- = \kappa_{d0}e^{-\eta_d b}$;
accordingly $S(\z) \in [S^-, S^+]$ with $S^\pm = K_p^\pm + K_d^\pm$.
Moreover, on $(a,b)$ the maps $K_p$ and $S$ are Lipschitz with constants
$L_p = \eta_p K_p^+$ and $L_S = \eta_p K_p^+ + \eta_d K_d^+$.
We first observe that the dynamics are bounded. Indeed, if $\gamma \le 1$
and $\varepsilon S^+ \le \gamma$, the box $[0,H]^N$ with
\begin{equation}
\label{eq.H}
H \;=\; \frac{\varepsilon K_p^+}{1 - \gamma + \varepsilon S^-}
\end{equation}
is forward-invariant for the MF-H-ESN dynamics under any input, so that
a trajectory started at $\h_0 = \mathbf{0}$ never leaves it
(Lemma~\ref{lem:invariant} in \ref{app:proofs}).

We are now in a position to state a sufficient condition for the ESP of
the MF-H-ESN. More precisely, the following
Proposition~\ref{th:esp} gathers the constants introduced above into a
single scalar $C$, whose two terms account respectively for the leak and
for the recurrent coupling, and characterizes the values of $C$ for which
the reservoir is a contraction.

\begin{proposition}[Echo state property of the MF-H-ESN]
\label{th:esp}
Let $\M$ be the structured operator of eq.~\ref{eq.operator}, assume
$\gamma \le 1$ and $\varepsilon S^+ \le \gamma$, let $H$ be as in
eq.~\ref{eq.H}, and let
\begin{equation}
\label{eq.esp_condition}
C \;=\; \max\big(|\gamma - \varepsilon S^-|, |\gamma - \varepsilon S^+|\big)
\;+\; \varepsilon\, \rho_M\, L_R \big( L_S H + L_p \big).
\end{equation}
If $C < 1$, then the MF-H-ESN satisfies the ESP.
\end{proposition}

The proof is reported in \ref{app:proofs}, and bounds the state
difference of two trajectories term by term. We can notice that
orthogonality is used at a single step, where
$\|\M(\h - \h')\|_2 = \|\h - \h'\|_2$ holds \emph{with equality}. As a
consequence, the recurrent contribution to $C$ is exactly $\rho_M$,
without any of the slack discussed in Remark~\ref{rem:tightness}.
Moreover, all the constants entering $C$ depend only on the endpoints
$a, b$, i.e., not on the state nor on the driving signal, so that the
contraction is uniform in the input.

\begin{remark}[Tightness, and comparison with dense reservoirs]
\label{rem:tightness}
Contraction arguments involve the spectral norm $\|\cdot\|_2$, whereas
the quantity that is tuned in reservoir practice is the spectral
radius. For the structured operator the two coincide, i.e.,
$\|\rho_M \M\|_2 = \rho_M$. For a dense random matrix rescaled to spectral
radius $\rho$, instead, $\|\W_h\|_2$ typically exceeds $\rho$ by a factor
close to $2$ (edge of the circular law vs.\ largest singular value), so
that norm-based sufficient conditions are loose by the same factor.
Orthogonality thus eliminates the slack between the tuned and the
contraction-relevant quantity. We note that this property is shared by any
orthogonal recurrence, including the SCR. The Hadamard
operator and the SCR therefore share the same stability, and differ in
their \emph{mixing}: a single application of $\M$ propagates information
from every coordinate to every other, whereas the SCR requires $N$ steps
for a full traversal.
This difference is measurable, and it grows with the reservoir size. The
measurements closing this section quantify it in terms of memory and
mixing, and the scaling experiments in terms of accuracy
(Section~\ref{sec:scaling}).
\end{remark}

\begin{remark}[Standard H-ESN]
In the case of the leaky-tanh H-ESN, the same argument gives the familiar
condition. Indeed, since $\tanh$ is 1-Lipschitz and
$\|\rho_M \M\|_2 = \rho_M$, any $\rho_M < 1$ guarantees the ESP, again
with no slack, in contrast with the dense case.
Conversely, since every eigenvalue of $\rho_M \M$ has modulus exactly
$\rho_M$, the linearization of the update at the origin is unstable for
$\rho_M > 1$, and the classical necessary condition \cite{jaeger2001echo}
excludes the ESP (for $\alpha = 1$ and admissible null input and bias).
For $\alpha=1$, the threshold $\rho_M = 1$ is therefore exact for the H-ESN: the gap
between the necessary and the sufficient condition, which for a dense
reservoir amounts to the factor discussed in Remark~\ref{rem:tightness},
closes entirely.
\end{remark}

\begin{remark}[On the conservativeness of the bound]
With the physics-related constants used in the experiments
($\kappa_{p0} = 10^{-4}$, $\eta_p = 10$, $\kappa_{d0} = 0.5$,
$\eta_d = 1$, $a = 0.35$, $b = 1.15$), the dominant constant is
$K_p^+ = \kappa_{p0} e^{\eta_p b} \approx 9.9$, so that, e.g., for
$\gamma = 0.95$, $\varepsilon = 10^{-2}$, $s = 1$, condition
$C < 1$ guarantees the ESP for $\rho_M \lesssim 0.09$.
As is typical of global Lipschitz arguments, the condition is sufficient
but far from necessary: the bound is dominated by the potentiation rate
at the upper end of the Rescale range, a regime that the driven dynamics
visit only sporadically, and empirically the models operate stably in the
entire explored hyper-parameter range. Sharper conditions could be
obtained by a local analysis around the operating regime, which we leave
for future work.
\end{remark}

\begin{remark}[Noise non-amplification]
\label{rem:noise}
We can extend the same tightness argument from the contraction of initial
conditions to the propagation of runtime perturbations. To this end, we
consider additive state noise injected at every step, i.e.,
$\h_t \mapsto \h_t + \sigma \boldsymbol{\xi}_t$, with
$\boldsymbol{\xi}_t$ i.i.d.\ standard Gaussian. 
To isolate the recurrent operator, consider the linear, non-leaky perturbation model $\delta\h_t=\rho_M\M\delta\h_{t-1}+\sigma\boldsymbol{\xi}_t$, corresponding to $\alpha=1$ and a unit activation derivative. For $\rho_M<1$, its stationary covariance is $\mathbf{P}=\sigma^2\sum_{k\geq0}\rho_M^{2k}\M^k(\M^k)^\top =\sigma^2(1-\rho_M^2)^{-1}\mathbf{I}$.
%
The value relies on the linearization, the independence of the realization
on $\M$ being orthogonal for every draw.
For a dense
random recurrence the same sum is governed by the singular values of
$\W_h^k$, which non-normality spreads about $\rho^k$ rather than
concentrating at it: the amplification is no longer uniform across
directions, and near the edge of stability, where the slowly decaying
terms dominate the sum, the upper tail yields a larger and markedly
realization-dependent noise gain. For the MF models, finally, the
per-step contraction is dominated by the effective leak
$\gamma - \varepsilon S$ rather than by $\rho_M$
(eq.~\ref{eq.esp_condition}), so the noise gain is essentially
flat in $\rho_M$. 
%
These qualitative predictions are empirically verified
in Section~\ref{sec:theory_diagnostics}. For both the Hadamard reservoirs, the
practical consequence is that the noise response is \emph{fixed at design
time}, i.e., it does not vary across realizations:
in the H-ESN it is set by the
single scalar $\rho_M$, through the exact orthogonality of the operator,
and in the MF-H-ESN by the neuron parameters $\gamma$ and $\varepsilon$,
whose effective leak dominates the contraction. A dense random recurrence
gives neither, since its gain varies from one realization to the next.
\end{remark}

\subsection{Spectral and mixing properties}
\label{sec:theory_mixing}
\begin{figure}[!htbp]
\centering
\includegraphics[width=\linewidth]{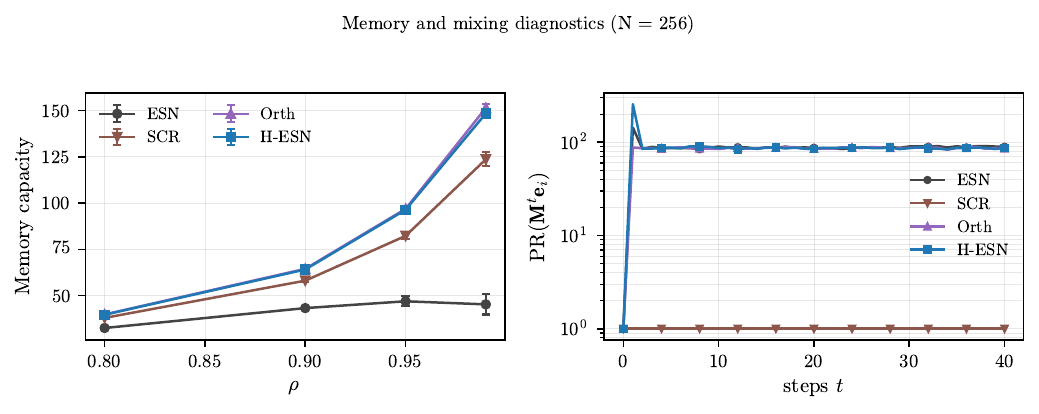}
\caption{Memory and mixing at $N = 256$ (mean $\pm$ std over
3 seeds). Left: memory capacity vs.\ the spectral radius $\rho$. The orthogonal
recurrences (SCR, dense orthogonal, Hadamard) dominate the dense random
ESN, and the Hadamard operator matches the dense orthogonal one. Right:
participation ratio (eq.~\ref{eq.pr}) of $\M^t \mathbf{e}_i$ vs.\ $t$
(log scale), at $\rho = 0.95$ and averaged over ten choices of the
impulsed unit $i$; $\mathrm{PR}$ does not depend on $\rho$. The SCR
translates the impulse from one unit to the next without ever spreading
it ($\mathrm{PR} = 1$), whereas the Hadamard operator attains the maximum
$\mathrm{PR} = N$ in one step.}
\label{fig:memory}
\end{figure}

We now move from stability to the \emph{quality} of the state
representation, i.e., to the way in which the structure of $\M$ shapes
the information that the reservoir retains. We make three observations,
and we support the first two with measurements on synthetic signals,
collected in Figure~\ref{fig:memory}. These compare four recurrent
couplings at $N = 256$, scaled to a common spectral radius $\rho$, i.e.,
the Hadamard operator $\M$, whose scaling $\rho_M$ plays exactly this
role, a dense random orthogonal matrix, the SCR, and the dense random
matrix of a standard ESN. The first three are orthogonal, and the last one
is not. The third observation concerns instead the ablated variants of the
operator, and we validate it together with the rest of the ablation, in
Section~\ref{sec:ablation}.

We first look at the spectrum. Since $\rho_M \M$ is a scaled orthogonal
matrix, \emph{all} its eigenvalues lie exactly on the circle of radius
$\rho_M$. A dense random reservoir rescaled to spectral radius $\rho$ has
instead eigenvalues spread across the whole disk of radius $\rho$
(circular law): most modes are strongly contractive and forget rapidly,
and memory is effectively carried by the few modes near the edge. In the
orthogonal case, instead, every direction of the state space retains
information at the same uniform rate $\rho_M$.
This is the spectral mechanism behind the near-optimal memory capacity of
orthogonal reservoirs \cite{white2004short,farkas2016memory}, and the
measurement agrees with it. We quantify the memory by the short-term
memory capacity (MC), measured with the standard protocol
\cite{white2004short,rodan2011minimum}, i.e., i.i.d.\ input in
$U(-0.8, 0.8)$, ridge readouts trained to reconstruct $\x_{t-k}$ for
$k \le 2N$, and $\mathrm{MC} = \sum_k r_k^2$ on held-out data ($\tanh$
reservoirs, $\omega_x = 0.1$, no bias, 3 seeds). At $\rho = 0.99$ it is
$148$ for the Hadamard operator and $151$ for the dense random orthogonal
matrix, against $45$ for the dense random reservoir of a standard ESN,
and the ordering holds over the whole range $\rho \in [0.80, 0.99]$
(Figure~\ref{fig:memory}, left). The structure therefore costs nothing in
memory, while the non-orthogonal recurrence retains about three times
less, saturating well below the others and degrading near the edge of
stability with an increasing variance across realizations.

We then turn to the speed at which the coupling spreads information,
which the spectrum does not determine.
We
can notice that the SCR shares
this spectral picture (the cycle is itself orthogonal, with eigenvalues on
the circle at the $N$-th roots of unity). 
Every entry of $\M$ equals $\pm 1/\sqrt{N}$, so a single application
of $\M$ propagates each state coordinate to every other coordinate; the
cyclic shift instead moves information by one position per step, requiring
$\Theta(N)$ steps for a full traversal of the state. This is directly
measurable on a coordinate impulse, i.e., by setting a single unit $i$ to
$1$ and all the others to $0$, and following the vector
$\M^t \mathbf{e}_i$ obtained after $t$ applications of the coupling alone.
To count how many units are active in it we use the participation ratio,
which for a generic vector $\mathbf{v} \in \R^N$ is
\begin{equation}
\label{eq.pr}
\mathrm{PR}(\mathbf{v}) \;=\;
\frac{\big(\sum_{j} v_j^2\big)^2}{\sum_{j} v_j^4},
\end{equation}
and which equals $1$ when a single coordinate carries the whole vector,
and $N$ when all the coordinates carry an equal share, so that it can be
read as an effective number of active units. Being homogeneous of degree
zero, it does not change if the state is rescaled, i.e., it measures how
the state is spread and not how large it is, and its value does not depend
on $\rho$, which we set to $0.95$. Measured on
$\mathbf{v} = \M^t\mathbf{e}_i$ and averaged over
ten choices of the impulsed unit, it confirms the two extremes
(Figure~\ref{fig:memory}, right): the Hadamard operator reaches
$\mathrm{PR} = N = 256$ after a single step, which is the largest value
the measure can take, and then settles at $\mathrm{PR} \approx N/3$
together with the dense couplings, whereas the SCR stays at
$\mathrm{PR} = 1$ at every horizon, since a cyclic shift translates the
impulse from one unit to the next without ever spreading it.

Finally, we can ask what the randomization elements $\mathbf{D}_1$, $\Pi$
and $\mathbf{D}_2$ contribute, and we find that they remove two
degeneracies. To this end, it is instructive to consider what happens
without them, i.e., when the recurrent operator is the bare transform
$\widehat\H_N$. We can notice that such an operator would already satisfy
both observations above, as it is orthogonal and it mixes the
whole state in a single step. Interestingly, however, it makes a poor
reservoir, and for two distinct reasons.
First, $\widehat\H_N$ is an involution, i.e., $\widehat\H_N^2 =
\mathbf{I}$, and applying it twice returns the state where it started.
The resulting reservoir alternates between two states instead of
exploring the state space, and its whole spectrum collapses onto the two
points $\pm 1$. Interposing the permutation $\Pi$ breaks the involution,
and spreads the eigenvalues over the circle.
Second, $\widehat\H_N$ cancels any drive that reaches all the units with
the same value, and concentrates it onto a single unit. Here a
permutation cannot help, as it maps a constant drive into a constant
drive, while the sign diagonals can (see \ref{app:proofs} for details).
This second effect is invisible to a linear probe, which can reconstruct
a delayed input whichever unit is carrying it. It emerges only under the
coordinatewise nonlinearity, where it determines whether the state
manifold stays rank one. Both degeneracies are measured empirically in
the ablation of Section~\ref{sec:ablation}.


Taken together, the results in Figure~\ref{fig:memory} show that the
Hadamard operator attains the memory of a dense orthogonal recurrence and
the one-step mixing of a dense random one, while retaining the cost and
the hardware complexity of a structured, multiplier-free operator. It
therefore occupies the middle ground described in
Section~\ref{sec:related}, between the maximally local topologies, which
are cheap but slow to mix, and the dense orthogonal recurrences, which
mix in one step but cost $O(N^2)$ to store and to apply.

\subsection{Noise response}
\label{sec:theory_diagnostics}
\begin{figure}[!htbp]
\centering
\includegraphics[width=0.55\linewidth]{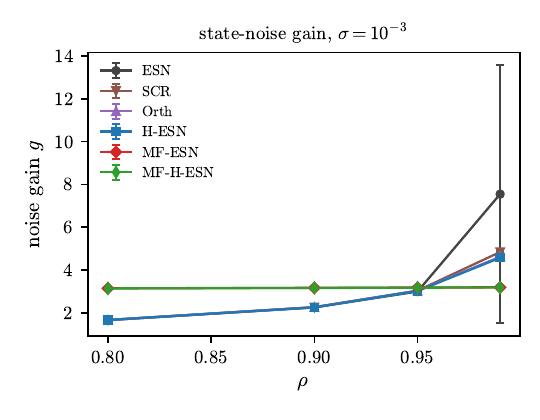}
\caption{Empirical state-noise gain on synthetic signals, vs.\ the
spectral radius $\rho$ ($N = 256$, mean $\pm$ std over 3 seeds). The
three orthogonal recurrences coincide, the dense random reservoir exceeds
them at the edge of stability with a large seed variance, and the MF
models show a flat gain, independent of $\rho$.}
\label{fig:robust-diag}
\end{figure}

We close the analysis by verifying the noise response predicted by
Remark~\ref{rem:noise}, injecting additive state noise of magnitude
$\sigma = 10^{-3}$ at every step 
in one of two copies driven by the same
input, and measuring the noise gain
$g = \|\tilde{\h}_t - \h_t\|_2/(\sigma\sqrt{N})$, i.e., the stationary
distance between the perturbed and the clean state, in units of the noise
injected at a single step.
%
Figure~\ref{fig:robust-diag} reports it as a function of $\rho$, and
confirms the three predictions of the Remark. The gain of the orthogonal
recurrences is set by $\rho$ alone, with negligible spread across
realizations, and follows $(1-\rho^2)^{-1/2}$ until the activation
saturates near the edge. The dense random reservoir has instead a gain
that varies from one realization to the next, which is the non-normality
effect discussed in the Remark. The MF models show a gain
that does not depend on $\rho$ at all, since their state decay is set by
the neuron kinetics rather than by the recurrence.

\section{Experimental evaluation}
\label{sec:experiments}

In this section we assess the Hadamard reservoirs empirically, on
time-series classification and regression benchmarks.
Section~\ref{sec:setup} describes the benchmarks, the baselines and the
model-selection protocol.
First, in Section~\ref{sec:fixedN} we compare all the models at a common
reservoir size of $N = 256$, which we take as the reference point of the
study.
Then, in Section~\ref{sec:scaling} we let the
reservoir size grow up to $N = 8192$, and analyze how the performance of
each model scales with it, on both the classification and the regression
benchmarks.
Finally, we look inside the operator and at its physical realizability:
Section~\ref{sec:ablation} ablates its components, and
Section~\ref{sec:robustness} quantifies the robustness of the models to
the perturbations that a physical implementation would encounter.

The experiments reported in this section were run on an NVIDIA A100 GPU.
The code used for our experiments is written in TensorFlow and
Scikit-learn, and is made publicly available
online.\footnote{\url{https://github.com/gallicch/Hadamard}}
Further hardware and software details are reported in \ref{app:protocol}.

\subsection{Experimental setup}
\label{sec:setup}
We evaluate the proposed architectures on twenty time-series
classification benchmarks and seven regression ones, drawn from the
UEA/UCR classification archive \cite{bagnall2018uea} and from the Monash
TSER archive \cite{tan2021tser}.\footnote{Classification datasets from
\url{https://timeseriesclassification.com}; regression datasets from
\url{http://tseregression.org}.} For classification, the original datasets
were split into training and test using a stratified 66\%--33\% division,
with a nested stratified 66\%--33\% split of the training portion for
model selection; input sequences were standardized to zero mean and unit
variance using statistics computed solely on the training data.
For regression, we used
the predefined train--test splits and an 80\%--20\% training--validation
split of the training portion.

We compare the proposed H-ESN and MF-H-ESN (with their simplified-input
variants, Section~\ref{sec:model_defs}) against the following baselines:
the conventional dense ESN of eq.~\ref{eq.esn_reservoir}; the dense
MF-ESN of eqs.~\ref{eq.mf_linear}--\ref{eq.mf_reservoir_h}; the canonical SCR
\cite{rodan2011minimum}, with identical cycle weights and input signs
determined by the decimal digits of $\pi$; a
dense random \emph{orthogonal} reservoir, obtained by QR decomposition of
a Gaussian matrix, which isolates the role of orthogonality from that of
structure; and a fully trained gated recurrent unit network (GRU)
\cite{cho2014learning}.
The GRU is a single recurrent layer with a linear or softmax head,
trained with Adam and early stopping on the validation split; its hidden
size, up to the same value $N = 256$ used for the reservoirs, and its
learning rate are selected on validation.

Reservoir hyper-parameters are selected by random search over 500
configurations on the validation split, and the readout is trained on the
reservoir state at the final time step by ridge regression with the
regularization chosen by leave-one-out on the training set. All the
baselines undergo the same model-selection protocol. Unless stated otherwise, results are
averages and standard deviations over 3 random initializations (the SCR
being deterministic, its repetitions coincide).
\ref{app:protocol} reports further details, including the search spaces and the training
details of the GRU baseline.

\subsection{Comparison at fixed reservoir size}
\label{sec:fixedN}
Tables~\ref{tab.results_classification} and \ref{tab.results_regression}
report test performance at $N = 256$, our reference reservoir size and
the starting point of the scaling analysis of
Section~\ref{sec:scaling}, for the classification and regression
benchmarks respectively; the number of classes of each classification dataset is
indicated in parentheses, and the best result per row is in bold.
We start from the classification benchmarks of
Table~\ref{tab.results_classification}, where the two simplified-input
Hadamard variants lead the comparison. They attain the two best mean
accuracies over the twenty tasks (last row of the table), and are the best
model on eleven of them.
The dense orthogonal baseline is competitive, but it is the best model
on a single dataset only. The SCR, equally orthogonal but with a
localized cycle coupling, collapses on Epilepsy ($0.75$ against $0.94$--$0.97$
for the other reservoirs), which is the clearest case in which global
state mixing matters; the comparison between the two anticipates the
analysis of Section~\ref{sec:ablation}.
We find the trained GRU competitive with the reservoirs on several
datasets, and the single best model on three of them. It falls clearly
behind, however, on the tasks with the least favorable data-to-class
ratio: on
Adiac (37 classes, about nine training sequences per class in our
experimental settings) it reaches only $0.33$ against $0.53$--$0.66$ for the reservoirs, and it
trails on BasicMotions. We attribute this to the small-sample regime of these benchmarks, i.e.,
to a known strength of RC. Fitting the recurrent weights of a gated
network by gradient descent is data-hungry, whereas the reservoir models
solve only a convex ridge problem for the readout.
We can also notice that the structured coupling preserves the accuracy of
the memristive-friendly models, i.e., MF-H-ESN matches or slightly
improves MF-ESN on most datasets, while MF-H-ESN$_{si}$ remains
competitive on several of them, with more variability due to its fully
binary input structure.

We then turn to the regression benchmarks of
Table~\ref{tab.results_regression}. Here the mean squared errors depend
on the scale of each target, so the rows are best read one at a time. The
dense memristive-friendly model is the best on three of the seven rows,
i.e., on BeijingPM10 and on two of the three FloodModeling tasks, while on
Covid3Month all the models land
within a few percent of each other. The Hadamard variants perform similarly to the dense models they replace,
with H-ESN close to the conventional ESN and MF-H-ESN close to MF-ESN
(the simplified-input variants deviate more, and do so on the
FloodModeling tasks). The GRU is
the best on three of the remaining four rows, and the SCR on
FloodModeling2, while a reservoir outperforms the GRU on the three
FloodModeling problems by factors of $2.5$ to $13$. On these tasks the GRU also shows a seed-to-seed
variability of a different order, which follows from its iterative
training, in contrast with the closed-form ridge readout of the
reservoir models.\footnote{The hyper-parameter grid of the GRU was widened, with
respect to the classification experiments, after observing systematic
plateaus at the constant predictor. See \ref{app:protocol}.}

Across both types of task, the Hadamard recurrence preserves the
performance of the standard ESN models, and occasionally improves it. We
read this as evidence that the structured orthogonal operator is a viable
substitute for a dense recurrent matrix in conventional reservoirs as
well.

\begin{sidewaystable}[p]
  \centering
  \footnotesize
  \setlength{\tabcolsep}{4pt}
  \begin{tabular}{l||rrr|r||rr||rrr}
    \textbf{Dataset (\#cl.)} &
    \textbf{ESN}&
    \textbf{SCR}&
    \textbf{Orth}&
    \textbf{GRU}&
    \textbf{H-ESN}&
    \textbf{H-ESN}$_{si}$&
    \textbf{MF-ESN}&
    \textbf{MF-H-ESN}&
    \textbf{MF-H-ESN}$_{si}$
    \\
    \hline
    Adiac (37) &
    $0.53_{\pm 0.05}$ & $0.50_{\pm 0.00}$ & $0.57_{\pm 0.04}$ & $0.33_{\pm 0.08}$ &
    $0.57_{\pm 0.05}$ & $0.55_{\pm 0.05}$ &
    $0.59_{\pm 0.01}$ & $0.60_{\pm 0.04}$ & $\mathbf{0.66}_{\pm 0.02}$ \\
    ArticularyWordRecognition (25) &
    $0.99_{\pm 0.00}$ & $0.99_{\pm 0.00}$ & $0.99_{\pm 0.00}$ & $0.85_{\pm 0.03}$ &
    $0.99_{\pm 0.00}$ & $0.99_{\pm 0.00}$ &
    $0.98_{\pm 0.00}$ & $0.99_{\pm 0.00}$ & $\mathbf{0.99}_{\pm 0.00}$ \\
    BasicMotions (4) &
    $\mathbf{1.00}_{\pm 0.00}$ & $1.00_{\pm 0.00}$ & $0.93_{\pm 0.06}$ & $0.78_{\pm 0.13}$ &
    $1.00_{\pm 0.00}$ & $1.00_{\pm 0.00}$ &
    $0.90_{\pm 0.02}$ & $1.00_{\pm 0.00}$ & $1.00_{\pm 0.00}$ \\
    ECG5000 (5) &
    $0.95_{\pm 0.00}$ & $0.95_{\pm 0.00}$ & $0.95_{\pm 0.00}$ & $\mathbf{0.95}_{\pm 0.00}$ &
    $0.95_{\pm 0.00}$ & $0.95_{\pm 0.00}$ &
    $0.95_{\pm 0.00}$ & $0.95_{\pm 0.00}$ & $0.95_{\pm 0.00}$ \\
    ERing (6) &
    $0.96_{\pm 0.00}$ & $0.96_{\pm 0.00}$ & $0.95_{\pm 0.01}$ & $0.83_{\pm 0.05}$ &
    $0.96_{\pm 0.00}$ & $\mathbf{0.98}_{\pm 0.00}$ &
    $0.93_{\pm 0.01}$ & $0.96_{\pm 0.01}$ & $0.98_{\pm 0.01}$ \\
    Earthquakes (2) &
    $0.81_{\pm 0.00}$ & $0.79_{\pm 0.00}$ & $0.81_{\pm 0.01}$ & $0.82_{\pm 0.01}$ &
    $0.81_{\pm 0.01}$ & $\mathbf{0.84}_{\pm 0.00}$ &
    $0.81_{\pm 0.00}$ & $0.81_{\pm 0.00}$ & $0.83_{\pm 0.00}$ \\
    Epilepsy (4) &
    $0.95_{\pm 0.01}$ & $0.75_{\pm 0.00}$ & $0.94_{\pm 0.01}$ & $0.83_{\pm 0.06}$ &
    $0.95_{\pm 0.01}$ & $\mathbf{0.97}_{\pm 0.00}$ &
    $0.95_{\pm 0.01}$ & $0.96_{\pm 0.00}$ & $0.96_{\pm 0.01}$ \\
    FaceDetection (2) &
    $0.65_{\pm 0.01}$ & $0.67_{\pm 0.00}$ & $0.65_{\pm 0.01}$ & $0.64_{\pm 0.01}$ &
    $0.66_{\pm 0.01}$ & $\mathbf{0.67}_{\pm 0.01}$ &
    $0.60_{\pm 0.01}$ & $0.60_{\pm 0.01}$ & $0.60_{\pm 0.00}$ \\
    FingerMovements (2) &
    $0.69_{\pm 0.00}$ & $0.68_{\pm 0.00}$ & $0.65_{\pm 0.01}$ & $0.64_{\pm 0.04}$ &
    $0.67_{\pm 0.01}$ & $\mathbf{0.71}_{\pm 0.01}$ &
    $0.65_{\pm 0.02}$ & $0.61_{\pm 0.00}$ & $0.67_{\pm 0.05}$ \\
    FordA (2) &
    $0.70_{\pm 0.01}$ & $0.72_{\pm 0.00}$ & $0.70_{\pm 0.00}$ & $0.72_{\pm 0.10}$ &
    $0.70_{\pm 0.01}$ & $0.71_{\pm 0.00}$ &
    $0.71_{\pm 0.01}$ & $\mathbf{0.72}_{\pm 0.00}$ & $0.71_{\pm 0.00}$ \\
    FordB (2) &
    $0.71_{\pm 0.02}$ & $0.75_{\pm 0.00}$ & $0.71_{\pm 0.00}$ & $\mathbf{0.79}_{\pm 0.10}$ &
    $0.74_{\pm 0.01}$ & $0.74_{\pm 0.01}$ &
    $0.74_{\pm 0.01}$ & $0.74_{\pm 0.00}$ & $0.74_{\pm 0.01}$ \\
    Handwriting (26) &
    $0.67_{\pm 0.02}$ & $0.58_{\pm 0.00}$ & $0.68_{\pm 0.01}$ & $0.57_{\pm 0.03}$ &
    $0.66_{\pm 0.01}$ & $0.62_{\pm 0.01}$ &
    $0.67_{\pm 0.00}$ & $0.65_{\pm 0.00}$ & $\mathbf{0.70}_{\pm 0.01}$ \\
    LSST (14) &
    $0.52_{\pm 0.00}$ & $0.54_{\pm 0.00}$ & $0.53_{\pm 0.00}$ & $\mathbf{0.62}_{\pm 0.01}$ &
    $0.52_{\pm 0.00}$ & $0.53_{\pm 0.00}$ &
    $0.59_{\pm 0.00}$ & $0.59_{\pm 0.00}$ & $0.58_{\pm 0.00}$ \\
    Libras (15) &
    $0.82_{\pm 0.01}$ & $0.82_{\pm 0.00}$ & $\mathbf{0.83}_{\pm 0.01}$ & $0.67_{\pm 0.03}$ &
    $0.76_{\pm 0.02}$ & $0.82_{\pm 0.02}$ &
    $0.83_{\pm 0.03}$ & $0.80_{\pm 0.01}$ & $0.79_{\pm 0.02}$ \\
    MindReading (5) &
    $0.64_{\pm 0.01}$ & $0.59_{\pm 0.00}$ & $0.62_{\pm 0.02}$ & $0.59_{\pm 0.01}$ &
    $0.63_{\pm 0.00}$ & $\mathbf{0.64}_{\pm 0.01}$ &
    $0.60_{\pm 0.01}$ & $0.58_{\pm 0.02}$ & $0.63_{\pm 0.01}$ \\
    NATOPS (6) &
    $0.95_{\pm 0.01}$ & $0.94_{\pm 0.00}$ & $0.94_{\pm 0.02}$ & $0.94_{\pm 0.01}$ &
    $0.95_{\pm 0.01}$ & $0.94_{\pm 0.00}$ &
    $\mathbf{0.96}_{\pm 0.01}$ & $0.95_{\pm 0.01}$ & $0.94_{\pm 0.00}$ \\
    PEMS-SF (7) &
    $0.80_{\pm 0.03}$ & $0.80_{\pm 0.00}$ & $0.78_{\pm 0.02}$ & $0.81_{\pm 0.00}$ &
    $0.85_{\pm 0.01}$ & $0.81_{\pm 0.01}$ &
    $0.84_{\pm 0.01}$ & $0.79_{\pm 0.03}$ & $\mathbf{0.88}_{\pm 0.01}$ \\
    PenDigits (10) &
    $\mathbf{0.99}_{\pm 0.00}$ & $0.99_{\pm 0.00}$ & $0.99_{\pm 0.00}$ & $0.99_{\pm 0.00}$ &
    $0.98_{\pm 0.00}$ & $0.99_{\pm 0.00}$ &
    $0.99_{\pm 0.00}$ & $0.99_{\pm 0.00}$ & $0.99_{\pm 0.00}$ \\
    PowerCons (2) &
    $0.98_{\pm 0.00}$ & $0.99_{\pm 0.00}$ & $0.97_{\pm 0.01}$ & $0.92_{\pm 0.01}$ &
    $1.00_{\pm 0.00}$ & $0.99_{\pm 0.02}$ &
    $\mathbf{1.00}_{\pm 0.00}$ & $0.99_{\pm 0.00}$ & $0.99_{\pm 0.01}$ \\
    RacketSports (4) &
    $0.89_{\pm 0.01}$ & $0.88_{\pm 0.00}$ & $0.89_{\pm 0.01}$ & $0.82_{\pm 0.03}$ &
    $0.87_{\pm 0.02}$ & $\mathbf{0.90}_{\pm 0.01}$ &
    $0.82_{\pm 0.01}$ & $0.86_{\pm 0.01}$ & $0.86_{\pm 0.02}$ \\
    \hline
    \textbf{Mean} &
    $0.810$ & $0.794$ & $0.804$ & $0.755$ &
    $0.811$ & $0.818$ &
    $0.805$ & $0.807$ & $\mathbf{0.823}$ \\
  \end{tabular}
  \caption{\footnotesize
  Test accuracy (higher is better; mean $\pm$ std over 3 runs) on the
  time-series classification benchmarks at $N=256$. The number of classes
  is in parentheses; best per row (by unrounded mean) in bold, and the
  last row averages each column over the twenty benchmarks. SCR has
  zero standard deviation by construction: cycle weights and
  $\pi$-digit input signs make it fully deterministic
  \cite{rodan2011minimum}. GRU is the only fully trained model (protocol in
  Section~\ref{sec:setup}). The subscript $_\mathrm{si}$ denotes the simplified-input variant defined in Section~\ref{sec:model_defs}.}
  \label{tab.results_classification}
\end{sidewaystable}

\begin{sidewaystable}[p]
  \centering
  \footnotesize
  \setlength{\tabcolsep}{4pt}
  \begin{tabular}{l||rrr|r||rr||rrr}
    \textbf{Dataset} &
    \textbf{ESN}&
    \textbf{SCR}&
    \textbf{Orth}&
    \textbf{GRU}&
    \textbf{H-ESN}&
    \textbf{H-ESN}$_{si}$&
    \textbf{MF-ESN}&
    \textbf{MF-H-ESN}&
    \textbf{MF-H-ESN}$_{si}$
    \\
    \hline
    AppliancesEnergy  &
    $7.83_{\pm 0.62}$ & $7.40_{\pm 0.00}$ & $7.34_{\pm 1.00}$ & $\mathbf{6.78}_{\pm 0.54}$ &
    $8.21_{\pm 0.33}$ & $9.28_{\pm 2.23}$ &
    $7.41_{\pm 0.88}$ & $7.29_{\pm 0.21}$ & $7.91_{\pm 0.40}$ \\
    BeijingPM10 $_{(\times 10^{3})}$ &
    $9.52_{\pm 0.35}$ & $9.87_{\pm 0.00}$ & $13.74_{\pm 3.57}$ & $9.67_{\pm 0.36}$ &
    $9.99_{\pm 0.54}$ & $9.90_{\pm 0.81}$ &
    $\mathbf{9.26}_{\pm 0.09}$ & $9.50_{\pm 0.19}$ & $9.58_{\pm 0.46}$ \\
    BeijingPM25 $_{(\times 10^{3})}$ &
    $4.47_{\pm 0.14}$ & $4.30_{\pm 0.00}$ & $4.36_{\pm 0.20}$ & $\mathbf{3.91}_{\pm 0.17}$ &
    $4.37_{\pm 0.27}$ & $4.12_{\pm 0.11}$ &
    $4.09_{\pm 0.04}$ & $4.21_{\pm 0.22}$ & $4.11_{\pm 0.21}$ \\
    Covid3M $_{(\times 10^{-3})}$ &
    $1.86_{\pm 0.00}$ & $1.85_{\pm 0.00}$ & $1.86_{\pm 0.00}$ & $\mathbf{1.81}_{\pm 0.07}$ &
    $1.86_{\pm 0.00}$ & $1.85_{\pm 0.00}$ &
    $1.84_{\pm 0.00}$ & $1.91_{\pm 0.00}$ & $1.83_{\pm 0.01}$ \\
    FloodM1 $_{(\times 10^{-5})}$ &
    $7.15_{\pm 0.48}$ & $10.61_{\pm 0.00}$ & $13.53_{\pm 3.94}$ & $13.77_{\pm 16.91}$ &
    $7.30_{\pm 0.25}$ & $10.10_{\pm 0.06}$ &
    $\mathbf{3.56}_{\pm 0.79}$ & $3.66_{\pm 0.56}$ & $5.88_{\pm 0.42}$ \\
    FloodM2 $_{(\times 10^{-5})}$ &
    $2.45_{\pm 0.16}$ & $\mathbf{1.87}_{\pm 0.00}$ & $2.31_{\pm 0.13}$ & $4.68_{\pm 0.58}$ &
    $2.05_{\pm 0.12}$ & $12.69_{\pm 5.69}$ &
    $2.76_{\pm 0.26}$ & $2.60_{\pm 0.21}$ & $3.68_{\pm 1.03}$ \\
    FloodM3 $_{(\times 10^{-5})}$ &
    $7.28_{\pm 0.11}$ & $7.72_{\pm 0.00}$ & $7.49_{\pm 0.23}$ & $51.87_{\pm 0.07}$ &
    $7.40_{\pm 0.13}$ & $7.55_{\pm 0.03}$ &
    $\mathbf{3.97}_{\pm 0.25}$ & $5.26_{\pm 0.46}$ & $7.67_{\pm 0.72}$ \\
  \end{tabular}
  \caption{\footnotesize
  Mean squared error (lower is better; mean $\pm$ std over 3 runs) on the
  time-series regression benchmarks at $N=256$. Covid3Months is
  abbreviated Covid3M, FloodModeling FloodM; best per row in bold. As in
  Table~\ref{tab.results_classification}, SCR is deterministic and has
  zero standard deviation by construction. The subscript $_\mathrm{si}$ denotes the simplified-input variant defined in Section~\ref{sec:model_defs}.}
  \label{tab.results_regression}
\end{sidewaystable}

\subsection{Scaling with the reservoir size}
\label{sec:scaling}

We next ask whether the structured models remain competitive as the
reservoir grows, i.e., in the regime where their $O(N \log N)$ compute and
their $O(N)$ parameters become decisive.
Hyper-parameters are selected once at $N = 256$, with the protocol of
Section~\ref{sec:setup}, and reused across sizes, following the common
practice of selecting the global parameters on a small reservoir and
then scaling it up, as the selected values are usually transferable to
bigger reservoirs \cite{lukovsevivcius2012practical}. Results are averaged over 3 random
instantiations, and per-dataset curves are reported in
\ref{app:scaling_all}.

\begin{figure}[!htbp]
\centering
\includegraphics[width=\linewidth]{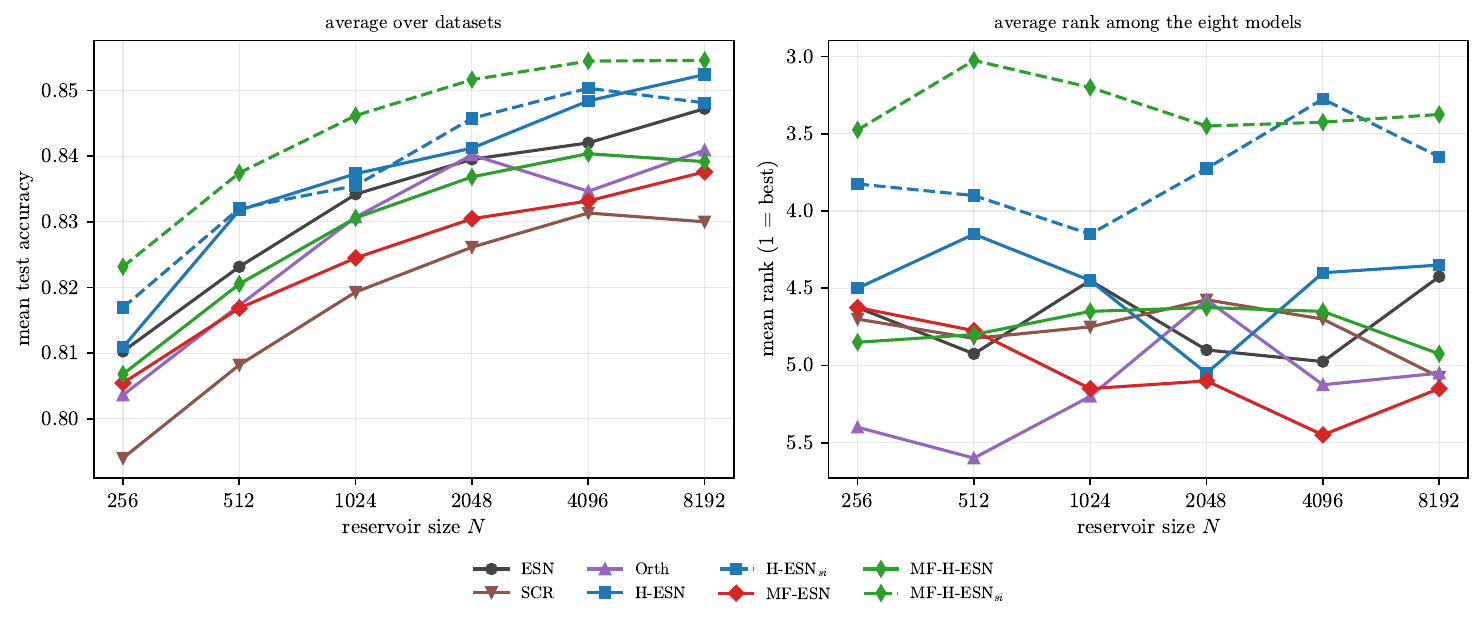}
\caption{Scaling with the reservoir size $N = 2^8,\ldots,2^{13}$ over the
twenty classification benchmarks (3 instantiations each). Left: test
accuracy averaged over datasets. Right: average rank among the eight
models, computed per dataset (rank 1 is best; the axis is inverted so
that better is up). Hyper-parameters are selected at $N=256$ and reused
across sizes for all models. The subscript $_\mathrm{si}$ denotes the simplified-input variant defined in Section~\ref{sec:model_defs}. Per-dataset curves: \ref{app:scaling_all}.}
\label{fig:scaling}
\end{figure}

We start from the classification benchmarks, whose test accuracy for
$N = 2^8, \ldots, 2^{13}$ is summarized in Figure~\ref{fig:scaling}, for
the eight reservoir models. The left panel averages accuracy across
benchmarks, and the right one averages the rank of each model, computed
benchmark by benchmark. We take the rank as the more robust of the two,
since averaging accuracies mixes tasks of different difficulty and number
of classes.

Three trends emerge. First, every model benefits from scale, by a comparable amount (between $3.1$ and $4.2$ accuracy points on average from $N = 256$ to $N = 8192$), so in absolute terms the structured
recurrence keeps pace with the dense one. Instead, scale changes the
ordering among the models.
A Hadamard-based model attains the best mean accuracy on twelve of the
twenty benchmarks at $N = 256$ and on thirteen at $N = 8192$, and the
average-rank panel shows the ordering to be stable across the whole
range, rather than the product of a single size.
Second, the two \emph{simplified-input} variants are consistently the
best-ranked models at every size, with MF-H-ESN$_{si}$ first throughout:
at $N = 8192$ it is $1.7$ points above the dense MF-ESN it simplifies
(better on 14 of 20 datasets) and $0.7$ above the conventional ESN. This
is worth emphasizing because that variant is also the most
hardware-frugal of the eight: a single binary input wire per unit, a
multiplier-free recurrence, and memristive-friendly neuron dynamics.
Third, the behavior is explained by mixing on top of orthogonality.
H-ESN stays at or above the dense orthogonal reservoir at every size, by
up to $1.5$ accuracy points, which is consistent with orthogonality being
the operative property. However, H-ESN delivers this property without ever
materializing the matrix, whereas
generating a dense orthogonal one by QR costs $O(N^3)$. Against the
equally orthogonal but local SCR, H-ESN gains $2.2$ accuracy points at
$N = 8192$, which reflects the global mixing anticipated in
Section~\ref{sec:theory} and dissected in the ablation below.

\begin{figure}[!htbp]
\centering
\includegraphics[width=\linewidth]{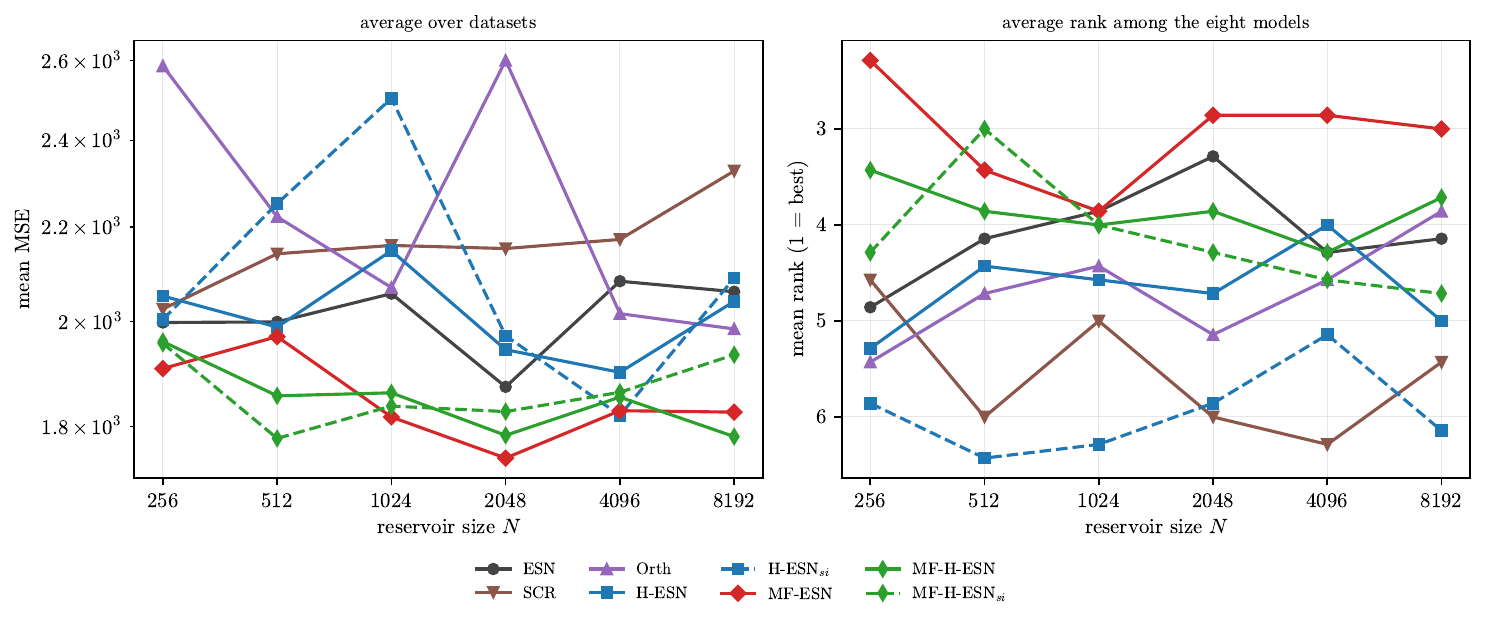}
\caption{Scaling with the reservoir size on the seven regression
benchmarks (3 instantiations each). Left: mean squared error averaged
over benchmarks, logarithmic scale. Right: average rank among the eight
models, computed per benchmark (rank 1 is best). Compare with
Figure~\ref{fig:scaling}: here the curves are nearly flat.}
\label{fig:scaling_reg}
\end{figure}

We then turn to the seven regression benchmarks of
Figure~\ref{fig:scaling_reg}, where the reservoir size plays a smaller
role than in classification. Averaged over the benchmarks, the mean
squared error is nearly flat over the whole range of sizes, against the
three to four accuracy points that every model gained on the
classification benchmarks. We read
this as a property of these tasks rather than of the models, i.e., the
targets are scalar and, on several of these benchmarks, weakly
determined by the input series, so that additional state dimensions have
little to encode, and whatever advantage a larger reservoir confers here
is already available at $N = 256$.

The ordering among the models is compressed accordingly, and two
observations can be drawn from it. First, on these tasks the
memristive-friendly neuron is what pays off, i.e., the dense MF-ESN is
the best-ranked model at five of the six sizes, and MF-H-ESN$_{si}$
leads at the remaining one. This is consistent with the fixed-size
comparison of Section~\ref{sec:fixedN}, where the memristive-friendly
models took the three FloodModeling benchmarks. Second, the Hadamard
reservoirs remain competitive at every scale, as a Hadamard variant is
among the three best-ranked models at each size, MF-H-ESN being the
second best at $N = 8192$. Its error stays within a few percent of the
corresponding dense model, MF-ESN, on most benchmarks and at every size
(\ref{app:scaling_all}). The structured recurrence therefore
keeps pace with the dense one here as well, in a regime in which scale
itself brings little to any model.

\begin{figure}[!htbp]
\centering
\includegraphics[width=\linewidth]{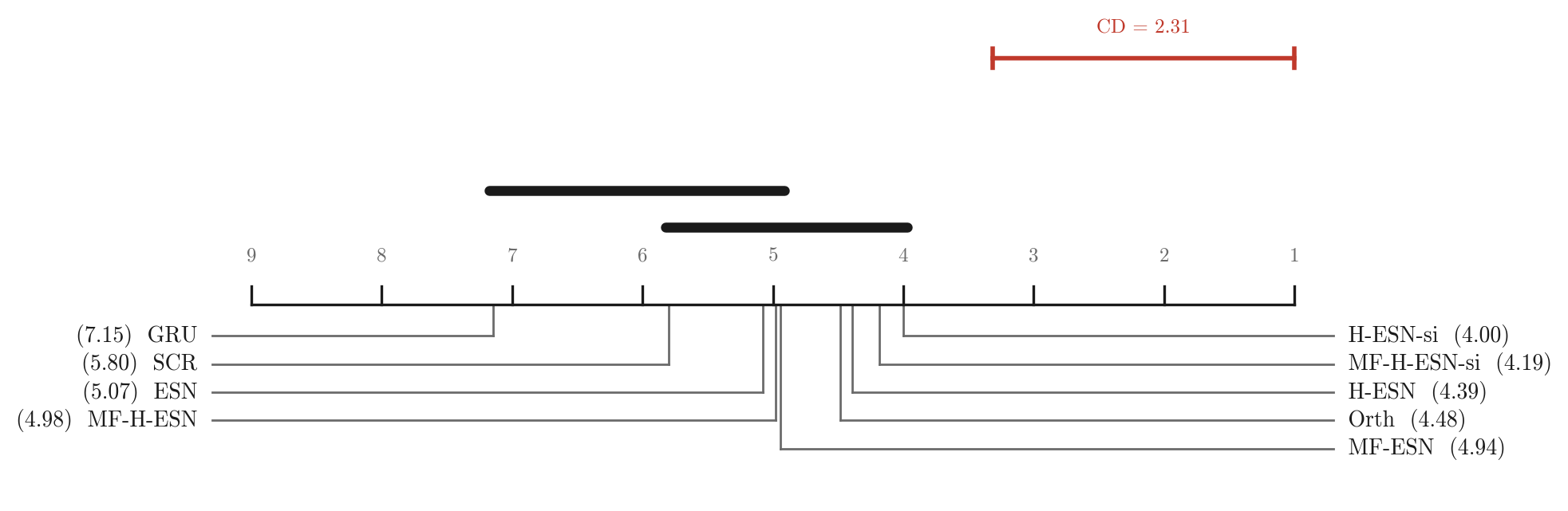}
\caption{Critical difference diagram over the twenty-seven benchmarks
(twenty classification, seven regression), with the reservoir size
selected on validation for the reservoir models and the hidden size
selected on validation for the GRU. Lower rank is better; models joined
by a bar are not significantly different at the $5\%$ level
(Friedman--Nemenyi, $\mathrm{CD} = 2.31$; Friedman test on the nine
models, $p = 5\cdot 10^{-4}$).}
\label{fig:cd}
\end{figure}

Across both types of task, Figure~\ref{fig:cd} puts these trends on a
statistical footing over all the twenty-seven benchmarks. There the
reservoir size is treated as one more
hyper-parameter selected on validation, i.e., the same treatment the GRU
receives for its hidden size, so that every model enters the comparison
at the capacity that validation selects for it. The two selections
follow the same protocol, over ranges that are comparable in the number
of trainable parameters, i.e., in the quantity that is actually fitted
to the data (\ref{app:params}).
Following the standard protocol for comparing models over multiple
datasets \cite{demsar2006statistical}, a Friedman test finds the
differences among the nine models to be statistically significant. The
Nemenyi post-hoc analysis then locates them, i.e., it separates the
trained GRU from the four best-ranked reservoirs, while the eight
reservoir models all lie within one critical distance of each
other.\footnote{A Friedman test restricted to the eight reservoir models
does not reject the null hypothesis of equal performance
($p = 0.20$).}
Three of the four Hadamard variants occupy the first three positions by
mean rank, and are statistically indistinguishable from the dense
baselines. These results confirm that the structured recurrence matches
dense reservoirs in accuracy, at a fraction of their cost.

\subsection{Ablation of the structured operator}
\label{sec:ablation}
\begin{figure}[!htbp]
\centering
\includegraphics[width=\linewidth]{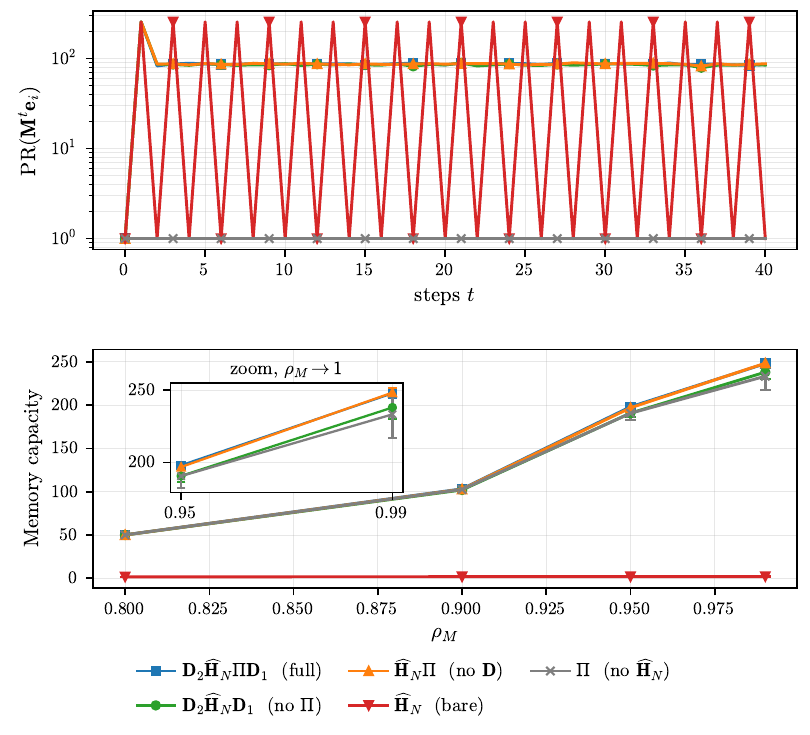}
\caption{The ablation variants as operators ($N = 256$). Top:
participation ratio (eq.~\ref{eq.pr}) of $\M^t\mathbf{e}_i$ at
$\rho_M = 0.95$, i.e., the number of active coordinates $t$ steps after a
coordinate impulse; the bare transform $\widehat\H_N$ alternates between
the whole state and a single unit, as its involutivity prescribes, and
the permutation-only control $\Pi$ never spreads the impulse. Bottom:
memory capacity of the linear reservoir vs.\ $\rho_M$, with an
inset zooming near the edge of stability, where removing the permutation
lowers the memory and raises its variance, while removing the sign
diagonals does not.}
\label{fig:ablation}
\end{figure}

\begin{figure}[!htbp]
\centering
\includegraphics[width=\linewidth]{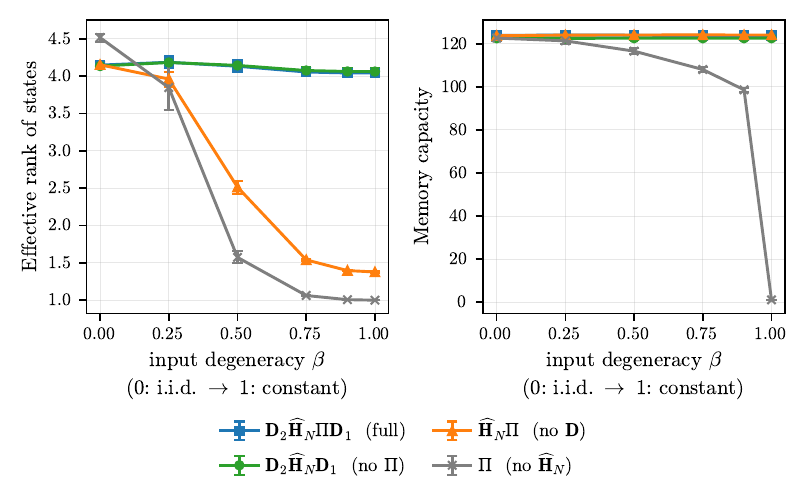}
\caption{Sensitivity to a degenerate input map ($N = 256$,
$\rho_M = 0.9$, $20$ seeds). The single column of the input weight matrix
$\W_x$ is interpolated at
fixed norm between i.i.d.\ entries ($\beta = 0$) and the constant vector
$\mathbf{1}$ ($\beta = 1$), along which every unit receives the same
drive. Left: the effective rank of the resulting nonlinear states, i.e.,
the participation ratio (eq.~\ref{eq.pr}) of their singular values, which
counts the populated directions of the state space. It collapses for the
operators without sign
diagonals ($\widehat\H_N\Pi$ and $\Pi$), while the full operator and
$\mathbf{D}_2\widehat\H_N\mathbf{D}_1$ (no permutation) keep it flat. Right:
the memory capacity is flat for every variant that retains the
Hadamard mixing, so the effect is purely nonlinear; only the mixing-free
control $\Pi$ degrades, and just at $\beta = 1$. Cf.\
\ref{app:proofs}.}
\label{fig:win}
\end{figure}

To isolate the contribution of each factor of the structured operator
$\M = \mathbf{D}_2\,\widehat\H_N\,\Pi\,\mathbf{D}_1$
(eq.~\ref{eq.operator}), we compare it against four reduced variants
obtained by switching components off while keeping $\widehat\H_N$ fixed:
$\mathbf{D}_2\widehat\H_N\mathbf{D}_1$ (no permutation), $\widehat\H_N\Pi$
(no sign diagonals), the bare transform $\widehat\H_N$, and the
permutation-only control $\Pi$ (mixing removed). All variants are rescaled
to the same spectral norm $\rho_M$, so they differ in structure alone and
not in contraction rate. We probe them at $N = 256$, from two
complementary points of view.
Figure~\ref{fig:ablation} characterizes each variant as an operator,
through the participation ratio of $\M^t\mathbf{e}_i$ (top,
eq.~\ref{eq.pr}) and the memory capacity as a function of $\rho_M$
(bottom). Figure~\ref{fig:win} then measures the
representational richness of the resulting nonlinear states, through
their effective rank, as the input map, i.e., the input weight matrix
$\W_x$ of eq.~\ref{eq.esn_reservoir} that injects the driving signal into
the reservoir, degrades from well-randomized to degenerate.

As anticipated in Section~\ref{sec:theory}, the bare transform
$\widehat\H_N$ is degenerate, and the top panel of
Figure~\ref{fig:ablation} shows its involutivity
($\widehat\H_N^2=\mathbf{I}$) directly in the mixing, i.e., a coordinate
impulse is spread over the whole state and then returned to a single
unit, alternating between the two at every step, whereas the variants
that combine $\widehat\H_N$ with a randomization element keep the impulse
spread at every horizon. The same degeneracy collapses both the memory
($\mathrm{MC}\approx 2$, bottom panel) and the diversity of the states
(effective rank $\approx 2$ out of $N$), which
confirms that some randomization is indispensable.
We find that the two remaining elements,
the permutation $\Pi$ and the sign diagonals, protect two \emph{different}
properties, which we take in turn.

We start from the memory curves of Figure~\ref{fig:ablation} (bottom),
where the permutation governs the behavior near the edge of stability. At
$\rho_M = 0.95$ the full operator reaches $\mathrm{MC} = 198.1 \pm 1.1$,
and the diagonal-free operator $\widehat\H_N\Pi$ matches it within seed
variability ($196.9 \pm 1.8$). Removing the permutation
($\mathbf{D}_2\widehat\H_N\mathbf{D}_1$) lowers it to $190.8 \pm 4.4$,
i.e., it costs about seven points of memory and multiplies the
seed-to-seed spread by four. The same pattern holds at $\rho_M = 0.99$,
with $247.9 \pm 3.1$ against $238.0 \pm 8.1$. The sign diagonals, in
contrast, play no role here: indeed, $\widehat\H_N\Pi$ reproduces the
full memory curve throughout.

We then turn to Figure~\ref{fig:win}, which asks how much the reservoir
relies on its input map being well randomized. We drive the models with a
scalar i.i.d.\ signal, so that $\W_x$ reduces to a single column, and we
interpolate that column, at
fixed norm, between two extremes: entries drawn i.i.d.\ ($\beta = 0$),
which is the standard initialization of Section~\ref{sec:esn}, and the
constant vector $\mathbf{1}$ ($\beta = 1$), along which every unit
receives exactly the same drive. On the resulting states we measure two
quantities. The first is their effective rank (left panel), i.e., the
participation ratio of eq.~\ref{eq.pr} evaluated on the singular values
of the collected states, which counts how many directions of the state
space are actually populated. The second is the memory capacity (right
panel).

Under a well-randomized input all the variants populate about four
directions. As the input concentrates on the constant vector, the
effective rank of the operators without sign diagonals collapses towards
$1$, i.e., their states end up
confined to a single direction, while the full operator and
$\mathbf{D}_2\widehat\H_N\mathbf{D}_1$ hold their rank. This confirms
what we anticipated in Section~\ref{sec:theory_mixing}, i.e., that a
permutation maps a constant drive into a constant drive, so that only the
sign diagonals can break it. The memory capacity, in the right panel, stays flat
across the whole sweep for every variant that retains the Hadamard
mixing, which confirms that the effect is purely
nonlinear, i.e., invisible to a linear probe.

Taken together, the ablation shows that the three structural elements of
$\M$ are individually necessary, and each for its own reason.
$\widehat\H_N$ supplies the one-step global mixing, the permutation $\Pi$
secures the memory near criticality, and the sign diagonals
$\mathbf{D}_1,\mathbf{D}_2$ preserve nonlinear state diversity under
adverse input conditions.

\subsection{Robustness to quantization and noise}
\label{sec:robustness}

We conclude the experimental analysis by subjecting the models to the
perturbations that a physical realization would encounter, and by
measuring what each of them costs in accuracy.
We consider two \emph{runtime} perturbations, i.e., additive state noise,
injected at every step with a standard deviation equal to a fraction
$\sigma$ of the root mean square (RMS) of the clean states, so that
$\sigma = 1\%$ perturbs a state by one hundredth of its typical
magnitude, and in-loop uniform quantization of the
state to $B$ bits, where $\h_t$ is quantized before re-entering the
recurrence. We also consider two \emph{permanent} perturbations of the
recurrent operator, i.e., a multiplicative device mismatch
$w \mapsto w (1 + \delta)$ with $\delta \sim \mathcal{N}(0, \sigma_d^2)$,
and faults in the discrete elements of the operator, i.e., random flips
of a fraction $p$ of the signs in $\mathbf{D}_1, \mathbf{D}_2$ and
faulty routing, obtained by drawing a fraction $p$ of the units and
permuting their destinations among themselves. The corrupted routing is
then again a permutation, i.e., the connections are misplaced but neither
duplicated nor lost. A fault that left a unit unconnected, or two units
driving the same destination, would take $\Pi$ out of the permutations and
$\M$ out of the orthogonal matrices, and we do not cover it here. Runtime perturbations are evaluated in two
regimes: a \emph{deploy} regime, with the readout trained on clean states
and the perturbation active only at test time, and an \emph{on-device}
regime, with the perturbation active also while the readout is trained,
as is standard practice for in-memory computing hardware
\cite{sebastian2020memory}. Permanent perturbations amount to a different
reservoir realization, and we evaluate them with the on-device readout
only. Hyper-parameters are reused from the model selection of
Section~\ref{sec:experiments} with no re-tuning, at $N = 256$ with $3$
reservoir realizations, on four of the classification benchmarks, chosen
to span short and long sequences, and univariate and multivariate inputs:
ECG5000, Epilepsy, RacketSports, and FordA, the last being the longest at
$T = 500$ steps.

\begin{figure}[!htbp]
\centering
\includegraphics[width=\linewidth]{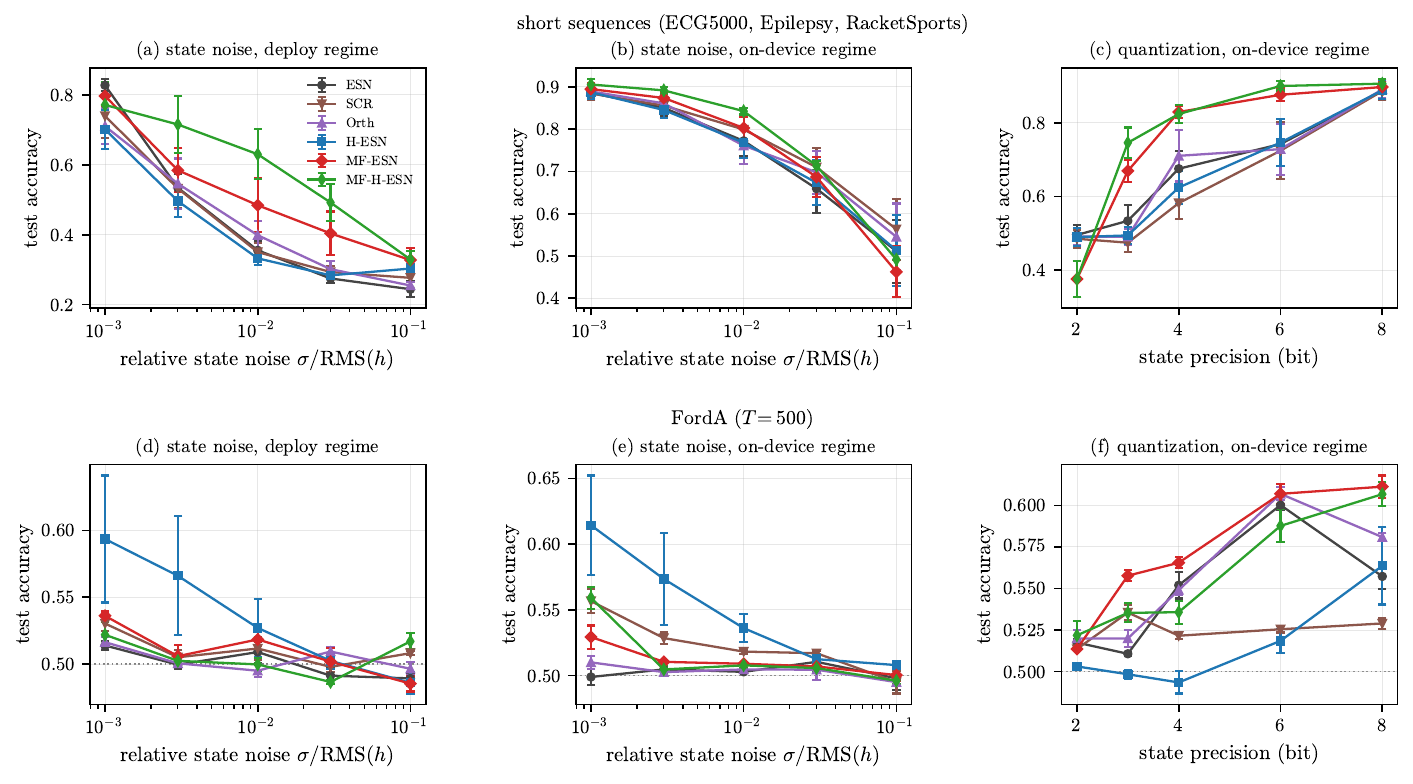}
\caption{Test accuracy under runtime perturbations at $N = 256$ (mean
$\pm$ standard error over 3 seeds; hyper-parameters from
Section~\ref{sec:experiments}, no re-tuning). Top row: mean over the three
short-sequence tasks (ECG5000, Epilepsy, RacketSports), where the
unperturbed accuracies lie between $0.91$ and $0.93$; bottom row: FordA
($T = 500$), where they lie between $0.61$ and $0.71$ and the dotted line
marks the $0.5$ level of this binary task. In the first two columns the perturbation \emph{grows} from
left to right, as the injected noise increases, in the deploy regime
(a, d) and in the on-device regime (b, e); in the third
column it \emph{shrinks} from left to right, as the state is quantized to
more bits (c, f). On the short tasks the on-device readout recovers most
of the noise cost and the MF-H-ESN is the most robust model; on FordA the
accumulated noise destroys the discriminative information for all models
and recalibration no longer helps.}
\label{fig:robust-data}
\end{figure}

We start from the accuracy cost of the runtime perturbations, reported
in Figure~\ref{fig:robust-data} with the three short tasks in the top row
and FordA in the bottom one. Three observations emerge. First, in the
deploy regime state noise is expensive for every model, i.e., on the short
tasks the accuracies fall from about $0.92$ to between $0.50$ and $0.72$
already at $\sigma = 0.3\%$, since a
ridge readout fitted on clean states amplifies any mismatch in the state
distribution. Second, the on-device readout recovers most of this loss on
the short tasks, where the memristive-friendly Hadamard model is
consistently the most robust, holding $0.84$ at $\sigma = 1\%$ against
between $0.76$ and $0.80$ for the leaky-tanh models, consistently with its
leak-dominated noise gain. FordA is the counterexample: over $500$ steps
the accumulated noise erases the discriminative information itself, so
the two regimes nearly coincide and every model sits on the $0.5$ level
of this binary task from $\sigma = 1\%$ on, the H-ESN being the last to
reach it. Third, in-loop quantization at $8$ bits is
essentially free for every model, which loses at most $4.4$ points of its
unperturbed accuracy with the on-device readout, and at $4$--$6$ bits the
MF models are the most robust on the short tasks.

Table~\ref{tab:device} completes the picture with the permanent
perturbations. Static variability turns out to be essentially free. The dense reservoirs
absorb even $20\%$ mismatch, since a jittered random matrix is
statistically still a random matrix and the recalibrated readout adapts
to it. The Hadamard models lose at most $2.3$ points at $20\%$
mismatch, whether the error acts on every entry of the materialized
operator or on the $2N\log_2 N$ nonzero entries of the butterfly stages,
and the loss is concentrated on FordA, where large gain errors visibly
break the exact orthogonality on which the guarantees of
Section~\ref{sec:theory} rest. Faults in the discrete elements are nearly
free throughout: corrupting $5\%$ of the signs costs at most $0.8$ points,
and rewiring $5\%$ of the routing costs less than half a point, which is
what the construction predicts, for the reason we discuss next.

Summarizing, every model tolerates these perturbations well once the
readout is trained on the device, the dense baselines included. What
distinguishes the Hadamard reservoir is where its margin comes from, i.e.,
from the construction rather than from the particular device. Its noise
response is fixed at design time by the exact orthogonality of the operator
and by the single tuned gain (Section~\ref{sec:theory_diagnostics}), so it
can be predicted under the model assumptions.
Little of its recurrence lives
on a continuous scale, and only what lives there can drift, i.e., one
programmable value and the $2N\log_2 N$ unit-gain coefficients of the
butterfly stages, against the $N^2$ conductances of a dense recurrence,
which is $4096$ quantities against $65536$ at $N = 256$. The rest of the
description is discrete, and the model class is closed under its
corruption, i.e., flipped signs return another pair of sign diagonals and a
rewired routing returns another permutation, so that the perturbed device
is another instance of the same family, with the same spectrum and the same
guarantees of Section~\ref{sec:theory}. What the on-device readout absorbs
is then a change of realization rather than a degradation of the dynamics,
which is why $5\%$ of the signs cost less than one accuracy point, and
$5\%$ of the routing less than half a point.

\begin{table}[!htbp]
\centering
\caption{Permanent device perturbations with on-device readout: change in
test accuracy (percentage points, mean over the four benchmarks and $3$
seeds) with respect to the unperturbed device. Crossbar mismatch applies
$w \mapsto w(1+\delta)$ to every entry of the materialized operator;
butterfly mismatch applies it to the $2N \log_2 N$ nonzero entries of
the FWHT stages; sign flips act on $\mathbf{D}_1, \mathbf{D}_2$, and
routing faults permute among themselves the destinations of a fraction of
the units, leaving $\Pi$ a permutation (both for the Hadamard models
only).}
\label{tab:device}
\footnotesize
\setlength{\tabcolsep}{3.2pt}
\begin{tabular}{l cccc c ccc c ccc}
\toprule
& \multicolumn{4}{c}{crossbar / butterfly mismatch $\sigma_d$}
& & \multicolumn{3}{c}{sign flips $p$}
& & \multicolumn{3}{c}{routing faults $p$} \\
\cmidrule{2-5} \cmidrule{7-9} \cmidrule{11-13}
Model & $1\%$ & $5\%$ & $10\%$ & $20\%$ & & $0.1\%$ & $1\%$ & $5\%$
& & $0.1\%$ & $1\%$ & $5\%$ \\
\midrule
ESN       & $-0.1$ & $+0.0$ & $+0.0$ & $-0.6$ & & --- & --- & --- & & --- & --- & --- \\
SCR       & $+0.1$ & $+0.1$ & $+0.5$ & $+0.1$ & & --- & --- & --- & & --- & --- & --- \\
Orth      & $-0.0$ & $+0.0$ & $+0.2$ & $-0.6$ & & --- & --- & --- & & --- & --- & --- \\
MF-ESN    & $-0.0$ & $-0.0$ & $-0.2$ & $-0.4$ & & --- & --- & --- & & --- & --- & --- \\
\midrule
H-ESN     & $-0.0$ & $-0.2$ & $-0.4$ & $-2.3$ & & $-0.4$ & $-0.4$ & $-0.8$
& & $+0.0$ & $-0.3$ & $+0.3$ \\
\quad (butterfly) & $+0.2$ & $-0.6$ & $-1.2$ & $-2.3$ & & & & & & & & \\
MF-H-ESN  & $-0.1$ & $+0.6$ & $-0.3$ & $-1.2$ & & $+0.1$ & $-0.0$ & $-0.4$
& & $+0.0$ & $-0.3$ & $-0.2$ \\
\quad (butterfly) & $+0.0$ & $-0.2$ & $+0.1$ & $-1.0$ & & & & & & & & \\
\bottomrule
\end{tabular}
\end{table}

\section{Discussion}
\label{sec:discussion}

Having introduced the models, analyzed them and measured them, we can now
put them side by side with the reservoir models that this work takes as
reference. Table~\ref{tab:summary} collects, in one place, the quantities
that the previous sections established separately.

\begin{table}[!htbp]
\centering
\caption{The proposed models against the reference ones, at a glance. The
columns are, in order, the conventional dense ESN, the Simple Cycle
Reservoir \cite{rodan2011minimum}, a dense random orthogonal recurrence,
the memristive-friendly MF-ESN \cite{pistolesi2025memristive}, which
shares the dense coupling of the ESN and differs in the neuron, and the
models introduced here, in their two input variants; in the last two rows,
where the two variants of a column differ, the entries refer to the
$\tanh$ and to the memristive-friendly neuron, in this order. The reservoir memory is the recurrent operator
together with the input pathway and the bias, in the form in which they
have to be resident to apply the recurrence, in single precision and on a
univariate task; the analog values are the entries of the same three
objects that a physical realization has to program one by one, and both
are detailed in \ref{app:params}. The readout, which is common to every
reservoir model and occupies $164$~kB in the same setting with five
classes, is excluded from both. Mixing, echo state condition and noise
gain are from Section~\ref{sec:theory}; the last two rows are the mean
rank among the eight models over the twenty classification and the seven
regression benchmarks at $N = 8192$, where rank $1$ is the best
(Section~\ref{sec:scaling}).}
\label{tab:summary}
\scriptsize
\setlength{\tabcolsep}{2.6pt}
\begin{tabular}{l ccccccc}
\toprule
& ESN & SCR & Orth & MF-ESN & H-ESN, & H-ESN$_{si}$, \\
& & & & & MF-H-ESN & MF-H-ESN$_{si}$ \\
\midrule
Operations per step & $N^2$ MAC & $N$ shift & $N^2$ MAC & $N^2$ MAC
& $N\log_2 N$ add & $N\log_2 N$ add \\
Construction cost & $O(N^2)$ & $O(1)$ & $O(N^3)$ & $O(N^2)$
& $O(1)$ & $O(1)$ \\
Reservoir memory, $N = 8192$ & $268$ MB & $2.1$ kB & $268$ MB & $268$ MB
& $80.9$ kB & $17.4$ kB \\
Analog values in the reservoir & $N^2{+}2N$ & $3$ & $N^2{+}2N$
& $N^2{+}2N$ & $2N{+}1$ & $3$ \\
\midrule
Impulse spread in one step & yes & no & yes & yes & yes & yes \\
Tight echo state condition & no & yes & yes & no & yes & yes \\
Noise gain fixed at design time & no & yes & yes & yes & yes & yes \\
Mean rank, classification & $4.4$ & $5.1$ & $5.1$ & $5.2$
& $4.4$ / $4.9$ & $3.7$ / $\mathbf{3.4}$ \\
Mean rank, regression & $4.1$ & $5.4$ & $3.9$ & $\mathbf{3.0}$
& $5.0$ / $3.7$ & $6.1$ / $4.7$ \\
\bottomrule
\end{tabular}
\end{table}

We start from the dense couplings. The conventional ESN, the dense
orthogonal reservoir and the MF-ESN
occupy the top half of the table, and they pay for it twice, i.e., with
$N^2$ analog values, which a physical substrate cannot afford, and with
$268$~MB of memory at $N = 8192$, which a digital one can afford only up to
a point. The structured operator obtains the
same one-step mixing at $N\log_2 N$ additions, one programmed gain, and a
description of tens of kilobytes, and it does so without loss on the
benchmarks, i.e., the best Hadamard variant is the best-ranked of the
eight models on the twenty classification benchmarks, and the dense
couplings are all behind it. Two qualifications complete the picture. First, the dense
\emph{orthogonal} recurrence is the closest competitor in the properties
that the theory isolates, since it is exactly orthogonal as well, but it
has to be generated by an $O(N^3)$ factorization and then stored and
applied densely, so what the structured operator removes there is the
cost, not the behavior. Second, on the regression benchmarks the MF-ESN
is the best-ranked model and MF-H-ESN follows it, within a few percent of
its error at every size (Section~\ref{sec:scaling}): on those tasks the
memristive-friendly neuron is what pays off, and the structured
recurrence keeps pace with the dense one rather than improving on it.

We then turn to the cycle. The SCR is the frugality reference of the
field, and it remains cheaper
than the operator proposed here, i.e., $N$ shifts per step against
$N\log_2 N$ additions, and a description of kilobytes against tens of
them. The comparison to make is the one with the simplified-input
variants, since four fifths of the $80.9$~kB of the plain H-ESN are the
real-valued input matrix and bias that it inherits from the standard ESN,
and only $15.4$~kB are the structured operator; once the input pathway is
binarized, as the SCR does with its own, the whole reservoir is $17.4$~kB.
Two observations then put the residual advantage in proportion. First, it does
not lie where the memristive scenario is decided: measured in analog
values, i.e., in quantities that a device has to program one by one and
that can then drift, the simplified-input Hadamard models are exactly at
the level of the cycle, i.e., three global scalars, independently of the
reservoir size. What separates the two is discrete and reproducible, i.e.,
signs and routing indices. Second, the absolute size of the gap is small
in the context of a complete model, i.e., $15$~kB at $N = 8192$, which is
less than a tenth of the ridge readout that both models carry anyway.
Against this, the cycle needs $N$ steps to spread a coordinate impulse
over the state, and it ranks $5.1$ on the classification benchmarks and
$5.4$ on the regression ones, against $3.4$ and $3.7$ for the best Hadamard
variant of each.
The structured operator therefore buys global mixing, and the accuracy
that follows from it, while staying in the same cost regime as the cycle.

It is finally worth measuring the structured recurrence per reservoir
unit. The operator stores two sign bits, one for each
diagonal, and one routing index of $\log_2 N$ bits, i.e., fifteen bits per
unit at $N = 8192$, and $N(2 + \log_2 N) = 15.4$~kB for the whole
recurrence. The transform itself stores nothing, since $\widehat\H_N$ is
fixed and is realized as a butterfly network; what it costs appears on the
other two axes, i.e., $N\log_2 N$ additions per step
(Section~\ref{sec:timing}) and $2N\log_2 N$ connections in a physical
realization (Section~\ref{sec:robustness}). A dense coupling stores $32N$
bits per unit in the same setting, i.e., about $33$~kB per unit against
two bytes.

\section{Conclusions}
\label{sec:conclusions}

We have studied reservoirs whose recurrent coupling is a structured
orthogonal operator built from sign diagonals, a permutation and a fast
Walsh-Hadamard transform, as a replacement for the dense random matrices
of conventional and memristive-friendly RC. The construction is
multiplier-free, costs $O(N \log N)$ operations and $O(N)$ parameters per
step, and never materializes a matrix.

Three findings stand out. First, 
we observe no systematic accuracy loss.
Across twenty classification and seven regression benchmarks, at reservoir
sizes up to $8192$, the Hadamard reservoirs track a dense random
orthogonal recurrence within one accuracy point at every size, while
avoiding the $O(N^3)$ factorization that generates such a matrix, and they are clearly
superior to the equally orthogonal SCR by a margin that grows with $N$.
The critical-difference analysis separates only the fully trained GRU from
the best-ranked reservoirs, so what these results support is parity at a
fraction of the cost. The advantage over the SCR comes from global mixing
on top of orthogonality, i.e., a coordinate impulse spreads over the
entire state in one step for the Hadamard operator, and never for the
cycle.

Second, the three components of the operator play distinct roles: the
permutation protects the memory near the edge of stability, the sign
diagonals protect the representational richness of the nonlinear states,
and the bare transform, being involutive, is degenerate on both counts. On
the classification benchmarks the two variants with a single binary input
wire per unit are the best-ranked models at every reservoir size, and they
are also the most hardware-frugal of those we considered.

Third, we have measured the computational advantage, and found that its
size depends on the device. The structured step overtakes the dense
product from $N \approx 1024$ on laptop-class CPUs and entry-level GPUs,
where it is about $50\times$ faster at $N = 32768$, and from
$N \approx 8192$ on an A100. The memory advantage, in contrast, does not
depend on the device and holds at every size. The recurrent operator is
described by $2N$ signs, $N$ routing indices and one gain, i.e.,
$15.4$~kB at $N = 8192$ against $268$~MB for a dense matrix, and with the
simplified input pathway the entire reservoir carries three analog values,
whatever its width. This is the decisive advantage in the memristive
scenario, and it is also what makes the device hard to corrupt, since
almost everything that is left to perturb is discrete.

On the theoretical side, the exact orthogonality of the operator removes
the slack that affects norm-based stability conditions for dense random
recurrences, and the resulting echo state condition is tight in the tuned
quantity. On the hardware side, the robustness study gives the numbers behind the
hardware-friendly design: the noise response is fixed at
design time, eight-bit state precision is essentially free, and 
in simulation, the models tolerate a $20\%$ mismatch,
corrupted signs or
faulty routing at a cost of about two accuracy points. The last of these
is a property of the construction rather than a measurement, since a
flipped sign or a rewired connection returns another valid instance of the
same model class, with the same spectrum and the same guarantees.

Section~\ref{sec:discussion} places these results next to the reference
models, in Table~\ref{tab:summary}. In one sentence, against the dense couplings the structured recurrence keeps the
representational properties and the accuracy while removing the $N^2$
analog weights that stand between a reservoir and a physical substrate,
and against the cycle it buys global mixing, and the accuracy that follows
from it, while remaining at the same three analog values.

Several directions remain open. A physical realization of the butterfly
operator on a memristive substrate is the natural continuation of this
work, together with an aging study, i.e., how long a readout calibrated
once remains valid as the analog components drift, where the reduced
analog surface of the structured implementation should pay off. A second
direction is architectural, since deep RC stacks several untrained layers
to obtain a hierarchy of timescales
\cite{gallicchio2017deep,gallicchio2017echo}, and the structured operator
can serve both within and between layers. A third one concerns the
\emph{temporal} axis. Deep state space models and linear recurrent units
\cite{gu2022s4,orvieto2023resurrecting}, and within RC the parallel echo
state networks of \cite{pinna2026paralesn}, face the same tension between
expressivity and cost, and resolve it by constraining the recurrence to be
linear and diagonal, so that it can be unrolled over the time steps by a
parallel scan. The operator studied here acts instead within a single
step, on the state dimension, so the two are composable rather than
alternative, and a structured multiplier-free mixing could replace the
dense projections that those models still carry. Finally, linear cycle reservoirs at the edge of
stability are known to represent the driving signal in the Fourier basis
\cite{tino2024linear}, and establishing which representation the linear
Hadamard reservoir realizes in the Walsh domain remains an open problem.

\section*{Acknowledgements}

This research has been partially funded by the EU-EIC EMERGE project (GA
101070918); by NEURONE, a project funded by the European Union --
Next Generation EU, M4C1 CUP I53D23003600006, under program PRIN 2022
(prj.\ code 20229JRTZA); and by the European Research Council (ERC)
under the European Union's Horizon Europe research and innovation
programme (ERC Starting Grant ``MEMBRAIN'', GA 101160604). Views and opinions expressed are however those of the
authors only and do not necessarily reflect those of the European Union
or the European Research Council. Neither the European Union nor the
granting authority can be held responsible for them.

\section*{Declaration of competing interest}

The authors declare that they have no known competing financial interests
or personal relationships that could have appeared to influence the work
reported in this paper.

\section*{Data availability}

The code used to reproduce the experiments is publicly available at
\url{https://github.com/gallicch/Hadamard}. The benchmark datasets are
publicly available from the UEA/UCR Time Series Classification Archive
and the Monash Time Series Extrinsic Regression Archive.

\appendix
\counterwithin*{figure}{section}
\counterwithin*{table}{section}
\renewcommand{\thefigure}{\Alph{section}.\arabic{figure}}
\renewcommand{\thetable}{\Alph{section}.\arabic{table}}

\section{Implementation of the structured recurrent step}
\label{app:implementation}

\begin{figure}[!htbp]
\centering
\includegraphics[width=0.72\linewidth]{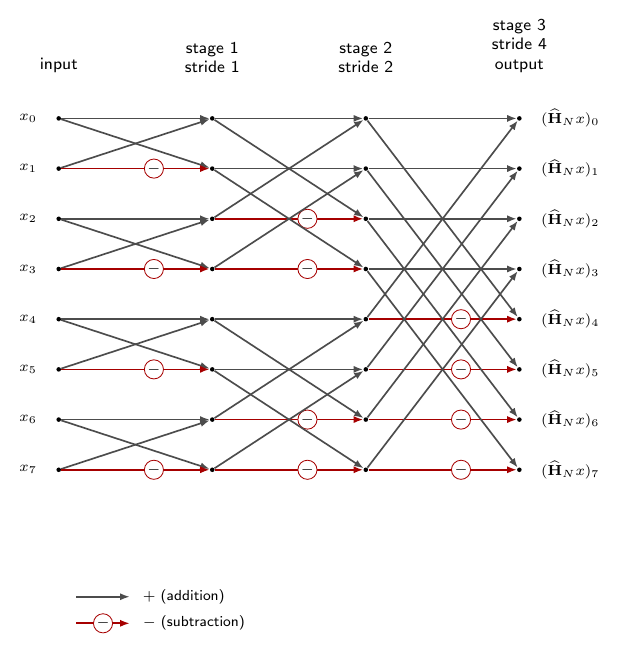}
\caption{Fast Walsh-Hadamard transform as a butterfly network, shown for
$N=8$ ($K=3$ stages, stride $1,2,4$). Each butterfly combines two inputs
into a sum (gray, $+$) and a difference (red, $-$). The coordinates need
no reordering, neither before nor after the stages, so the network
computes exactly the transform $\widehat\H_N$ of
Section~\ref{sec:operator_def}. In general, the $K=\log_2 N$ stages of
$N/2$ butterflies each amount to $N\log_2 N$ additions/subtractions and no
multiplications, against $N^2$ multiply-accumulates for the corresponding
dense matrix-vector product: at $N=4096$, $4.9\times 10^4$ additions
against $1.68\times 10^7$ multiply-accumulates, a $341\times$ reduction.}
\label{fig:butterfly}
\end{figure}

We detail here the three realizations of the recurrent step
$\h \mapsto \rho_M \M \h$ compared in Section~\ref{sec:implementation}.
The first is a pure TensorFlow FWHT based on the radix-2 butterfly
decomposition (Figure~\ref{fig:butterfly}), which executes the $\log_2 N$
stages as separate elementwise passes, and therefore launches $\log_2 N$
kernels per step. The second is the same computation compiled with XLA
(\texttt{jit\_compile}), which fuses the stages into a small number of
specialized GPU kernels. The third is a dedicated CUDA kernel, exposed as a
custom TensorFlow operator, which for $N \le 8192$ executes the whole
transform in a single shared-memory kernel, with one thread block per batch
row and $\log_2 N$ butterfly stages separated by block-level
synchronization, and for larger $N$ falls back to a multi-stage
global-memory implementation. It registers no gradient, since none is
back-propagated through the reservoir.

Tables~\ref{tab:crossover-cpu}--\ref{tab:crossover-a100} report the
measurements summarized in Figure~\ref{fig:crossover}, for the three
platforms separately. Software and hardware versions are declared in
\ref{app:protocol}.

\begin{table}[!htbp]
\centering
\footnotesize
\begin{tabular}{r|r|rr|rr|rr}
$N$ & dense [ms] & \multicolumn{2}{c|}{FWHT (TF)} & \multicolumn{2}{c|}{FWHT (XLA)} & \multicolumn{2}{c}{FWHT (C++ op)} \\
 & & [ms] & speedup & [ms] & speedup & [ms] & speedup \\ \hline
256   & 0.16   & 0.74  & 0.21$\times$  & 0.31  & 0.51$\times$  & 0.27  & 0.58$\times$ \\
512   & 0.38   & 1.16  & 0.32$\times$  & 0.39  & 0.97$\times$  & 0.42  & 0.90$\times$ \\
1024  & 1.19   & 2.23  & 0.53$\times$  & 0.56  & 2.13$\times$  & 0.72  & 1.64$\times$ \\
2048  & 3.28   & 4.30  & 0.76$\times$  & 0.98  & 3.35$\times$  & 1.47  & 2.23$\times$ \\
4096  & 11.43  & 10.09 & 1.13$\times$  & 2.00  & 5.72$\times$  & 2.85  & 4.01$\times$ \\
8192  & 43.20  & 19.27 & 2.24$\times$  & 3.82  & 11.29$\times$ & 5.72  & 7.55$\times$ \\
16384 & 173.60 & 41.38 & 4.20$\times$  & 8.26  & 21.02$\times$ & 9.66  & 17.96$\times$ \\
32768 & 830.63 & 65.47 & 12.69$\times$ & 16.70 & 49.74$\times$ & 22.82 & 36.40$\times$ \\
\end{tabular}
\caption{Time per recurrent step [ms] and speedup vs.\ dense matmul
(batch 256, fp32) on a laptop-class Apple~M3~Pro CPU.
The crossover occurs
at $N \approx 1024$ with operator fusion (XLA and C++ custom op) and at
$N \approx 4096$ for the unfused FWHT; the speedup profile closely mirrors
that of the T4 GPU (Table~\ref{tab:crossover}).}
\label{tab:crossover-cpu}
\end{table}

\begin{table}[!htbp]
\centering
\footnotesize
\begin{tabular}{r|r|rr|rr|rr}
$N$ & dense [ms] & \multicolumn{2}{c|}{FWHT (TF)} & \multicolumn{2}{c|}{FWHT (XLA)} & \multicolumn{2}{c}{FWHT (CUDA op)} \\
 & & [ms] & speedup & [ms] & speedup & [ms] & speedup \\ \hline
256   & 0.18   & 0.71  & 0.25$\times$  & 0.34  & 0.53$\times$  & 0.29  & 0.62$\times$ \\
512   & 0.38   & 1.18  & 0.32$\times$  & 0.44  & 0.87$\times$  & 0.43  & 0.88$\times$ \\
1024  & 1.16   & 2.12  & 0.55$\times$  & 0.63  & 1.84$\times$  & 0.74  & 1.57$\times$ \\
2048  & 3.28   & 4.16  & 0.79$\times$  & 1.10  & 3.00$\times$  & 1.57  & 2.09$\times$ \\
4096  & 11.74  & 10.50 & 1.12$\times$  & 2.31  & 5.09$\times$  & 3.08  & 3.81$\times$ \\
8192  & 48.63  & 20.90 & 2.33$\times$  & 4.56  & 10.67$\times$ & 6.49  & 7.49$\times$ \\
16384 & 195.54 & 33.71 & 5.80$\times$  & 9.05  & 21.62$\times$ & 10.23 & 19.12$\times$ \\
32768 & 822.16 & 75.43 & 10.90$\times$ & 16.63 & 49.43$\times$ & 22.42 & 36.66$\times$ \\
\end{tabular}
\caption{Time per recurrent step [ms] and speedup vs.\ dense matmul
(batch 256, fp32, NVIDIA T4).}
\label{tab:crossover}
\end{table}

\begin{table}[!htbp]
\centering
\footnotesize
\begin{tabular}{r|r|rr|rr|rr}
$N$ & dense [ms] & \multicolumn{2}{c|}{FWHT (TF)} & \multicolumn{2}{c|}{FWHT (XLA)} & \multicolumn{2}{c}{FWHT (CUDA op)} \\
 & & [ms] & speedup & [ms] & speedup & [ms] & speedup \\ \hline
256   & 0.20 & 0.52 & 0.39$\times$ & 0.22 & 0.90$\times$ & 0.30 & 0.66$\times$ \\
512   & 0.20 & 0.61 & 0.32$\times$ & 0.23 & 0.87$\times$ & 0.34 & 0.58$\times$ \\
1024  & 0.23 & 0.61 & 0.38$\times$ & 0.24 & 0.98$\times$ & 0.28 & 0.82$\times$ \\
2048  & 0.22 & 0.65 & 0.34$\times$ & 0.23 & 0.94$\times$ & 0.30 & 0.74$\times$ \\
4096  & 0.23 & 0.68 & 0.34$\times$ & 0.24 & 0.97$\times$ & 0.30 & 0.78$\times$ \\
8192  & 0.35 & 0.77 & 0.46$\times$ & 0.30 & 1.16$\times$ & 0.36 & 0.99$\times$ \\
16384 & 1.38 & 1.51 & 0.92$\times$ & 0.39 & 3.53$\times$ & 0.47 & 2.93$\times$ \\
32768 & 5.20 & 4.69 & 1.11$\times$ & 2.00 & 2.61$\times$ & 2.22 & 2.34$\times$ \\
\end{tabular}
\caption{Time per recurrent step [ms] and speedup vs.\ dense matmul
(batch 256, fp32, NVIDIA A100). On datacenter-class
hardware the dense baseline is substantially faster than on the T4
(Table~\ref{tab:crossover}), moving the crossover point from
$N\approx 1024$ to $N\approx 8192$; below that size all implementations
sit on a common kernel-launch latency floor and the structured step is
never slower than the dense one (XLA column).}
\label{tab:crossover-a100}
\end{table}

\section{Memory and parameter accounting}
\label{app:params}

We detail here the counts of Section~\ref{sec:memory}, separating two
questions that are easily conflated, i.e., how much of a model has to be
stored, and how much of it has to be trained.

We count the recurrent operator together with the input matrix and the
bias, and we keep the readout apart, since it is the same for every
reservoir model. Real-valued parameters are counted at $32$ bits, a sign at
one bit, an index into $N$ coordinates at $\log_2 N$ bits, and an index
into $d$ input channels at $\lceil\log_2 d\rceil$ bits. The global scalars
of each model, i.e., the recurrent scaling, the leaking rate and the input
and bias scalings, are a dozen numbers at most and are invisible in the
totals, so we leave them out of the stored description; we do count them
among the analog values, where they are the three entries of the
simplified-input models. The constants of the memristive-friendly kinetics
belong to the neuron rather than to the coupling, and are excluded from
both counts. Wiring is not memory, so the butterfly interconnect and the
cycle of the SCR appear among the analog connections only.
Table~\ref{tab:memory} collects the resulting formulas.

\begin{table}[!htbp]
\centering
\caption{Memory footprint of the reservoir description. Left: bits
required to store the recurrent operator and the input pathway, with
real-valued parameters at $32$ bits, signs at one bit, and indices at
$\log_2 N$ and $\lceil\log_2 d\rceil$ bits; the few global scalars per
model are negligible here and are omitted, but they are counted among the
analog values, which they exhaust for the simplified-input models. Middle:
quantities that a physical realization has to instantiate, i.e., analog
values to be programmed individually, and analog connections to carry
them; the butterfly interconnect of the FWHT and the cycle of the SCR are
wiring, and are counted here only. Right: resulting total for the
reservoir description on a univariate task ($d = 1$), at two reservoir
sizes. The readout, which is common to all the reservoir models, is listed
separately for a $C$-class task, with $C = 5$.}
\label{tab:memory}
\scriptsize
\setlength{\tabcolsep}{3.5pt}
\begin{tabular}{l rr rr rr}
\toprule
& \multicolumn{2}{c}{stored description (bits)}
& \multicolumn{2}{c}{analog quantities}
& \multicolumn{2}{c}{total, $d = 1$} \\
\cmidrule(lr){2-3} \cmidrule(lr){4-5} \cmidrule(lr){6-7}
Model & recurrent & input, bias & values & connections
& $N = 1024$ & $N = 8192$ \\
\midrule
ESN & $32N^2$ & $32N(d{+}1)$ & $N^2{+}N(d{+}1)$ & $N^2{+}Nd$
& $4.2$ MB & $268.5$ MB \\
SCR & $32$ & $N(d{+}1)$ & $3$ & $N{+}Nd$
& $260$ B & $2.1$ kB \\
Orth & $32N^2$ & $32N(d{+}1)$ & $N^2{+}N(d{+}1)$ & $N^2{+}Nd$
& $4.2$ MB & $268.5$ MB \\
MF-ESN & $32N^2$ & $32N(d{+}1)$ & $N^2{+}N(d{+}1)$ & $N^2{+}Nd$
& $4.2$ MB & $268.5$ MB \\
\midrule
H-ESN & $N(2{+}\log_2 N)$ & $32N(d{+}1)$ & $1{+}N(d{+}1)$
& $2N\log_2 N{+}Nd$ & $9.7$ kB & $80.9$ kB \\
H-ESN$_{si}$ & $N(2{+}\log_2 N)$ & $N(2{+}\lceil\log_2 d\rceil)$ & $3$
& $2N\log_2 N{+}N$ & $1.8$ kB & $17.4$ kB \\
MF-H-ESN & $N(2{+}\log_2 N)$ & $32N(d{+}1)$ & $1{+}N(d{+}1)$
& $2N\log_2 N{+}Nd$ & $9.7$ kB & $80.9$ kB \\
MF-H-ESN$_{si}$ & $N(2{+}\log_2 N)$ & $N(2{+}\lceil\log_2 d\rceil)$ & $3$
& $2N\log_2 N{+}N$ & $1.8$ kB & $17.4$ kB \\
\midrule
readout & \multicolumn{2}{c}{$32(N{+}1)C$}
& \multicolumn{2}{c}{$(N{+}1)C$} & $20.5$ kB & $163.9$ kB \\
\bottomrule
\end{tabular}
\end{table}

Once the recurrence is structured, the description of a Hadamard
reservoir is dominated by its input pathway, already on a univariate task,
i.e., of the $80.9$~kB of the H-ESN at $N = 8192$ only $15.4$~kB are the
operator, and the remaining $65.5$~kB are $\W_x$ and $\b_h$, real-valued
as in a standard ESN. The simplified pathway replaces that term with $2N$
signs and brings the whole reservoir to $17.4$~kB, of which the operator
is now the larger part. On the multivariate benchmarks the effect is more
pronounced, i.e., on MindReading ($d = 204$) at $N = 8192$ the input
matrix costs $6.7$~MB in the dense pathway and $10$~kB in the simplified
one.

The ridge readout maps the $N$-dimensional state to the class scores,
i.e., $(N+1)\,C$ weights for a $C$-class task. On a five-class task it
occupies $20.5$~kB at $N = 1024$ and $164$~kB at $N = 8192$, which is
nine times the entire H-ESN$_{si}$ reservoir and less than a thousandth of
the dense ESN.

Storage, however, is not what is fitted to the data. The reservoir models
train the readout alone, whereas in the GRU every weight is trained, i.e.,
$3h(d + h + 2)$ in the recurrent cell plus $(h+1)C$ in the head, with $h$
the hidden size selected on validation; at $h = 256$, the largest size we
select, the recurrent cell alone holds $796$~kB of real-valued weights on a
univariate task, more than the entire H-ESN reservoir at $N = 8192$,
readout included. At $N = 256$ the trained network optimizes more
parameters than the reservoir readout on every one of the twenty
benchmarks, by a median factor of $26$ and by up to $602$ on FaceDetection,
and it has to fit them by back-propagation, which is the size of the
data-efficiency gap invoked in Section~\ref{sec:fixedN}. At $N = 8192$ the
median ratio is $0.83$ instead, and the GRU
is the larger model on nine of the twenty benchmarks, i.e., measured in
state dimension the two families are not comparable, and measured in
trainable parameters they are. The two views differ because enlarging a
reservoir costs memory and inference time while leaving the demand for
gradients and data unchanged.

\section{Proofs}
\label{app:proofs}
This appendix collects the proofs of the results stated in
Section~\ref{sec:theory}, together with two results that are used there
but not reproduced in the main text: the invariant-set lemma and the
input degeneracy of the diagonal-free operator. Throughout,
the notation is that of eq.~\ref{eq.phi} and of the bounds on
$K_p, K_d, S$ established there.

\begin{lemma}[Invariant set]
\label{lem:invariant}
Assume $\gamma \le 1$ and $\varepsilon S^+ \le \gamma$, and let $H$ be
as in eq.~\ref{eq.H}. Then $[0, H]^N$ is forward-invariant for the
MF-H-ESN dynamics: if $\h_{t-1} \in [0,H]^N$ then $\h_t \in [0,H]^N$,
for any input.
\end{lemma}

\paragraph{Proof of Lemma~\ref{lem:invariant}}
Componentwise, $h_{t,i} = (\gamma - \varepsilon S(z_{t,i}))\, h_{t-1,i} +
\varepsilon K_p(z_{t,i})$, and the coefficient of $h_{t-1,i}$ lies in
$[\gamma - \varepsilon S^+,\, \gamma - \varepsilon S^-] \subseteq
[0,\, \gamma - \varepsilon S^-]$ by assumption. Hence
$h_{t,i} \ge \varepsilon K_p^- > 0$ and
$h_{t,i} \le (\gamma - \varepsilon S^-) H + \varepsilon K_p^+ \le H$,
where the last inequality is eq.~\ref{eq.H}.
\qed

\paragraph{Proof of Proposition~\ref{th:esp} (echo state property)}
Let $\h, \h' \in [0,H]^N$ and let $\x$ be an arbitrary input. By
Lemma~\ref{lem:invariant} the set $[0,H]^N$ is forward-invariant,
and starting from $\h_0 = \mathbf{0}$ every trajectory remains in it, so
the estimates below may be taken on that set, which, being closed and bounded, is complete. We write $\z = R(\rho_M \M \h + \W_x \x + \b_h)$ and
$\z' = R(\rho_M \M \h' + \W_x \x + \b_h)$ for the two pre-activations.
Adding and subtracting $(\gamma \mathbf{1} - \varepsilon S(\z)) \odot \h'$
gives the exact decomposition
\begin{equation}
\label{eq.esp_decomposition}
\Phi(\h,\x) - \Phi(\h',\x)
= (\gamma \mathbf{1} - \varepsilon S(\z)) \odot (\h - \h')
- \varepsilon (S(\z) - S(\z')) \odot \h'
+ \varepsilon (K_p(\z) - K_p(\z')),
\end{equation}
which we bound term by term in the Euclidean norm.
\emph{(i)} Since $\z$ takes values in $(a,b)$ componentwise, each entry of
the diagonal factor lies in $[\gamma - \varepsilon S^+,
\gamma - \varepsilon S^-]$, so the first term is at most
$\max(|\gamma - \varepsilon S^-|, |\gamma - \varepsilon S^+|)\,
\|\h - \h'\|_2$.
\emph{(ii)} Using $\|\mathbf{u} \odot \mathbf{v}\|_2 \le
\|\mathbf{u}\|_\infty \|\mathbf{v}\|_2$ together with
$\h' \in [0,H]^N$, hence $\|\h'\|_\infty \le H$, and the Lipschitz
property of $S$ on $(a,b)$, the second term is at most
$\varepsilon H L_S \|\z - \z'\|_2$.
\emph{(iii)} Likewise the third term is at most
$\varepsilon L_p \|\z - \z'\|_2$.
It remains to bound the pre-activations. The input and bias contributions
are identical in $\z$ and $\z'$ and cancel in the difference, so, $R$
being $L_R$-Lipschitz componentwise,
\begin{equation}
\|\z - \z'\|_2 \;\le\; L_R\, \rho_M \|\M(\h - \h')\|_2
\;=\; L_R\, \rho_M \|\h - \h'\|_2,
\end{equation}
where the last step holds \emph{with equality of norms} because $\M$ is exactly orthogonal. We can notice that this is the only point at which the
structure of the operator is used, and it is what removes the slack
discussed in Remark~\ref{rem:tightness}. Collecting the three bounds yields
$\|\Phi(\h,\x) - \Phi(\h',\x)\|_2 \le C \|\h - \h'\|_2$ with $C$ as in
eq.~\ref{eq.esp_condition}.
Finally, $C$ is built only from $\varepsilon, \gamma, \rho_M$ and the
constants $K_p^\pm, K_d^\pm, L_p, L_S, L_R$, which depend on the
endpoints $a, b$ of the range of $R$ and not on the state or the driving
signal: the contraction is therefore \emph{uniform in the input}. If $C < 1$, $\Phi(\cdot, \x)$ is thus a uniform contraction of a complete
state space, and the ESP follows by standard arguments
\cite{jaeger2007optimization,manjunath2013echo,ceni2020echo}. For every
left-infinite input the state sequence is unique and independent of the
initial conditions, and any two trajectories driven by the same input
converge at rate $C^t$.
\qed

We close with the input degeneracy invoked in the third observation of
Section~\ref{sec:theory_mixing}. Let $\mathbf{1}\in\R^{N}$ be the
constant direction, along which every unit receives the same drive, and
let $\mathbf{e}_1$ be the first canonical basis vector. The first row of
$\H_N$ is $\mathbf{1}^{\!\top}$ and the remaining rows are orthogonal to
it, so $\H_N\mathbf{1}=N\mathbf{e}_1$ and
$\widehat\H_N\mathbf{1}=\sqrt{N}\,\mathbf{e}_1$. Since
$\Pi\mathbf{1}=\mathbf{1}$, this yields, for every permutation $\Pi$,
\begin{equation}
\label{eq.dc-collapse}
\widehat\H_N\,\Pi\,\mathbf{1}=\sqrt{N}\,\mathbf{e}_1,
\qquad
\mathbf{D}_2\,\widehat\H_N\,\Pi\,\mathbf{D}_1\,\mathbf{1}
=\mathbf{D}_2\,\widehat\H_N\big(\Pi\mathbf{d}_1\big),
\end{equation}
the second identity using $\mathbf{D}_1\mathbf{1}=\mathbf{d}_1$, with
$\mathbf{d}_1\in\{\pm1\}^{N}$ the diagonal of $\mathbf{D}_1$, and the
definition of $\M$. The diagonal-free operator therefore concentrates the
constant direction onto a single unit, whatever the permutation. The full
operator spreads it instead: $\|\widehat\H_N(\Pi\mathbf{d}_1)\| =
\|\mathbf{d}_1\| = \sqrt{N}$ by orthogonality, and each entry of
$\widehat\H_N(\Pi\mathbf{d}_1)$ is $1/\sqrt{N}$ times a sum of $N$
independent signs, hence has zero mean and unit variance, so the energy
$N$ is spread evenly across the $N$ units in expectation rather than
concentrated on one.

\section{Experimental protocol: search spaces and baseline training}
\label{app:protocol}

We use twenty classification and seven regression tasks, drawn from the
UEA/UCR and Monash TSER archives. The datasets were chosen, before inspecting any result on them,
to obtain a heterogeneous collection on which to probe the models. They
span from $2$ to $37$ classes, from $8$ to $500$ time steps, from
univariate to $963$-dimensional inputs, and from a few hundred to some
ten thousand sequences, and they include both sensor and biomedical
recordings. Datasets whose sequence length or cardinality would have
made the full model-selection protocol impractical, or too small for a
reliable estimate, were left out.

All experiments use TensorFlow, and datasets are loaded through the aeon
toolkit \cite{middlehurst2024aeon}. The model experiments of
Section~\ref{sec:experiments} were run on an NVIDIA A100 GPU with
TensorFlow 2.21. The wall-clock analysis of
Section~\ref{sec:implementation} spans three platforms by design, with
TensorFlow 2.12 on the Apple~M3~Pro CPU and 2.21 on both GPUs (NVIDIA T4
and A100); its timings are medians over repeated executions after
warm-up, in single precision, with $256$ states advanced per call.

The random search of Section~\ref{sec:setup} samples 500 configurations
from the following ranges:
$\rho, \rho_M \in \{0.7, 0.8, \ldots, 1.2\}$;
$\omega_x, \omega_b \in \{0.01, 0.1, 1.0, 2.0, 5.0\}$;
$\alpha \in \{1, 0.1, 0.01, 0.001\}$; and, for the MF models,
$\varepsilon \in \{0.1, 0.01, 0.001\}$,
$\gamma \in \{0.5, 0.8, 0.95, 1.0\}$, $s \in \{1, 5\}$. The
physics-related constants are $\kappa_{p0} = 10^{-4}$,
$\kappa_{d0} = 0.5$, $\eta_p = 10$, $\eta_d = 1$, following
\cite{milano2022connectome}. The ridge regularization is selected in
$\{10^{-5}, \ldots, 10^{2}\}$.

The GRU is trained with Adam \cite{kingma2015adam}, batch size $32$,
for up to $500$ epochs, with early stopping (patience $50$) on validation accuracy for
classification and on validation loss for regression; the selected
configuration is retrained and evaluated over 3 seeds. The hidden size is
selected in $\{32, 64, 128, 256\}$ and the learning rate in
$\{10^{-3}, 3\!\cdot\!10^{-4}\}$.
On the regression benchmarks the learning-rate grid is extended upwards
to $\{10^{-2}, 3\!\cdot\!10^{-3}, 10^{-3}, 3\!\cdot\!10^{-4}\}$. The
reason is that with the classification grid the network converged to the
constant predictor on every regression benchmark. A sweep over learning
rate and batch size showed that this is a property of the learning-rate
range rather than of the architecture or of the sequence length: no configuration with $\eta \le 10^{-3}$ escapes the plateau at any batch size, whereas $\eta = 10^{-2}$ fits the training set to less than one
hundredth of the error of the constant predictor. The wider grid is used
for the GRU only and only in regression: the classification results are
those obtained with the original grid, under which no such pathology was
observed: we verified explicitly that the GRU exceeds the majority-class
rate on all twenty classification datasets, although by a narrow margin
on Earthquakes ($0.82$ against $0.80$), whose majority class already
accounts for four fifths of the data.
Targets are standardized when training the GRU and the predictions are
mapped back before computing the error, so that all reported errors are
in the original scale; this affects the baseline only marginally and is
not what determines whether it escapes the plateau.

The comparison at fixed reservoir size of Section~\ref{sec:fixedN} is set
up to be fair to the trained baseline, and we record here the two points
where the two families are treated differently. The reservoir size is
fixed to $N = 256$ for all the reservoir models, whereas the GRU hidden
size is itself selected on validation, up to that same value: the two are
therefore matched in state dimension, and if anything the choice favors
the trained network, since smaller hidden sizes are typically selected
because larger recurrent layers overfit these small datasets. The GRU
keeps instead the internal train/validation split at assessment time, as
early stopping requires a held-out set, whereas the readout of the
reservoir models is refit on the full training set after model
selection.

The critical-difference analysis of Section~\ref{sec:scaling} treats the
reservoir size as one more hyper-parameter, selected on the same
validation split already used for the others. The additional selection
is over six values only, against the 500 configurations already searched
there, and it requires no further model selection, i.e., the validation
accuracies per size are obtained by reloading the selected
hyper-parameters and refitting. The GRU receives the same treatment for
its hidden size, so that the two selections follow the same protocol.
The reservoirs may reach $N = 8192$ and the GRU hidden size is capped at
$256$, and the two ranges remain comparable in the number of trainable
parameters, which is the quantity fitted to the data, as we quantify in
\ref{app:params}.

\section{Per-dataset scaling curves}
\label{app:scaling_all}

\begin{figure}[p]
\centering
\includegraphics[width=\linewidth,height=0.92\textheight,keepaspectratio]{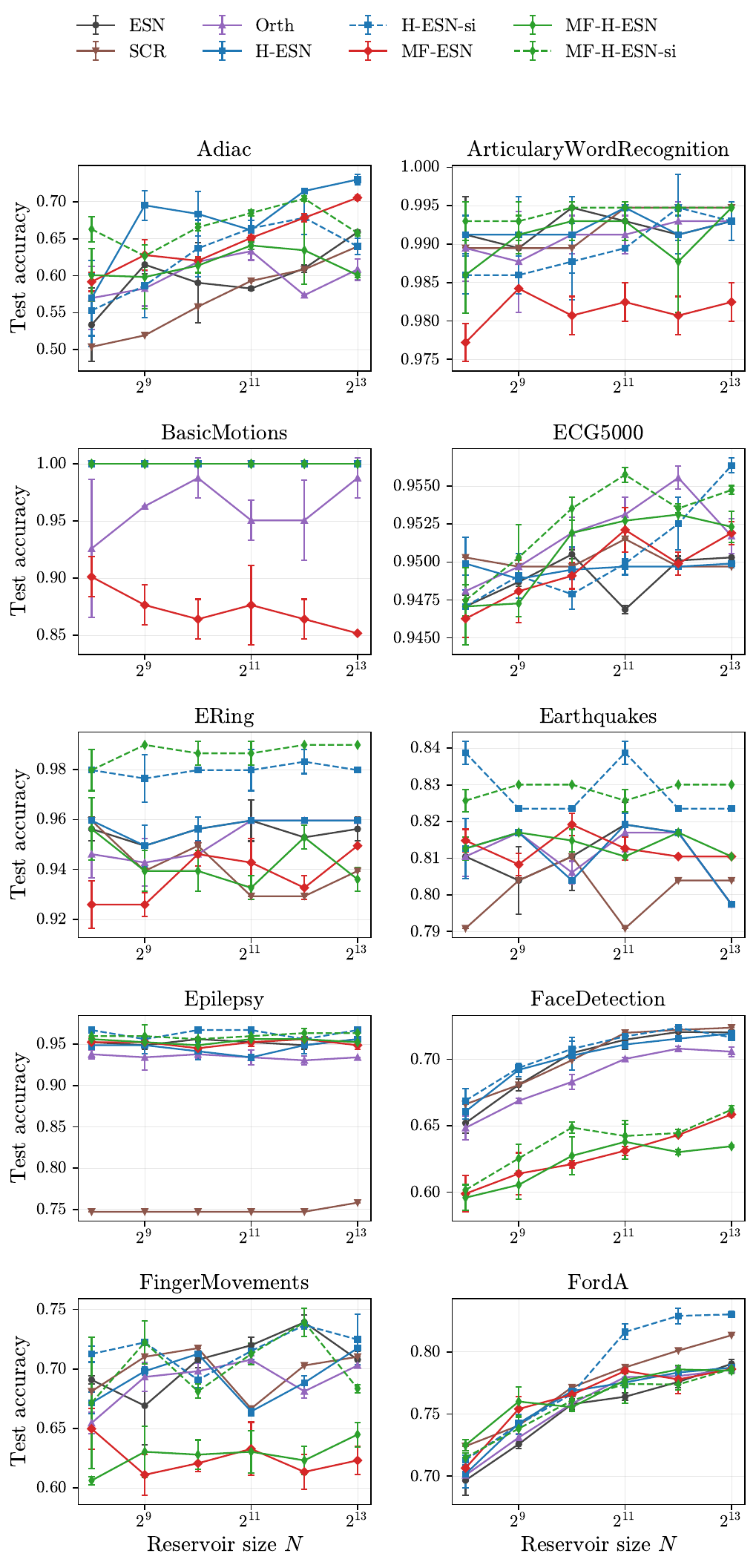}
\caption{Test accuracy (mean $\pm$ std over 3 instantiations) versus
reservoir size, one panel per classification benchmark (first ten,
in alphabetical order; the remaining ten in
Figure~\ref{fig:scaling_all_b}).}
\label{fig:scaling_all}
\end{figure}

\begin{figure}[p]
\centering
\includegraphics[width=\linewidth,height=0.92\textheight,keepaspectratio]{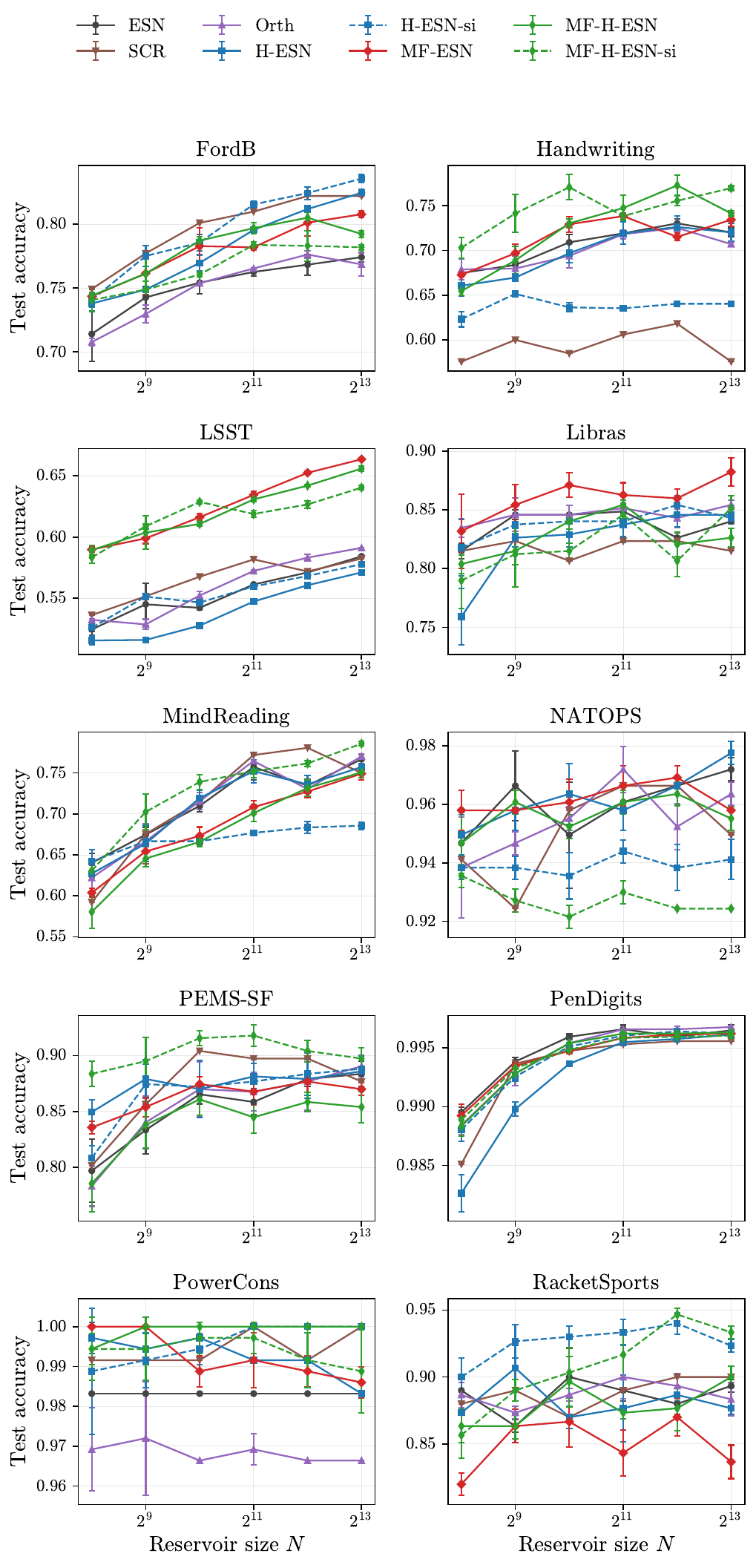}
\caption{Test accuracy versus reservoir size, remaining ten
classification benchmarks. Continuation of
Figure~\ref{fig:scaling_all}.}
\label{fig:scaling_all_b}
\end{figure}

Figures~\ref{fig:scaling_all} and~\ref{fig:scaling_all_b} report, for
each of the twenty classification benchmarks separately, the test accuracy of the eight
reservoir models as a function of $N$; Figure~\ref{fig:scaling} in the
main text is the aggregate of these curves. The per-dataset view makes
visible what the averages hide. On several benchmarks accuracy saturates
well before $N = 8192$, on a few it degrades at the largest sizes, and
the spread between models is much wider on the hardest tasks than the
average suggests.

\begin{figure}[p]
\centering
\includegraphics[width=\linewidth,height=0.92\textheight,keepaspectratio]{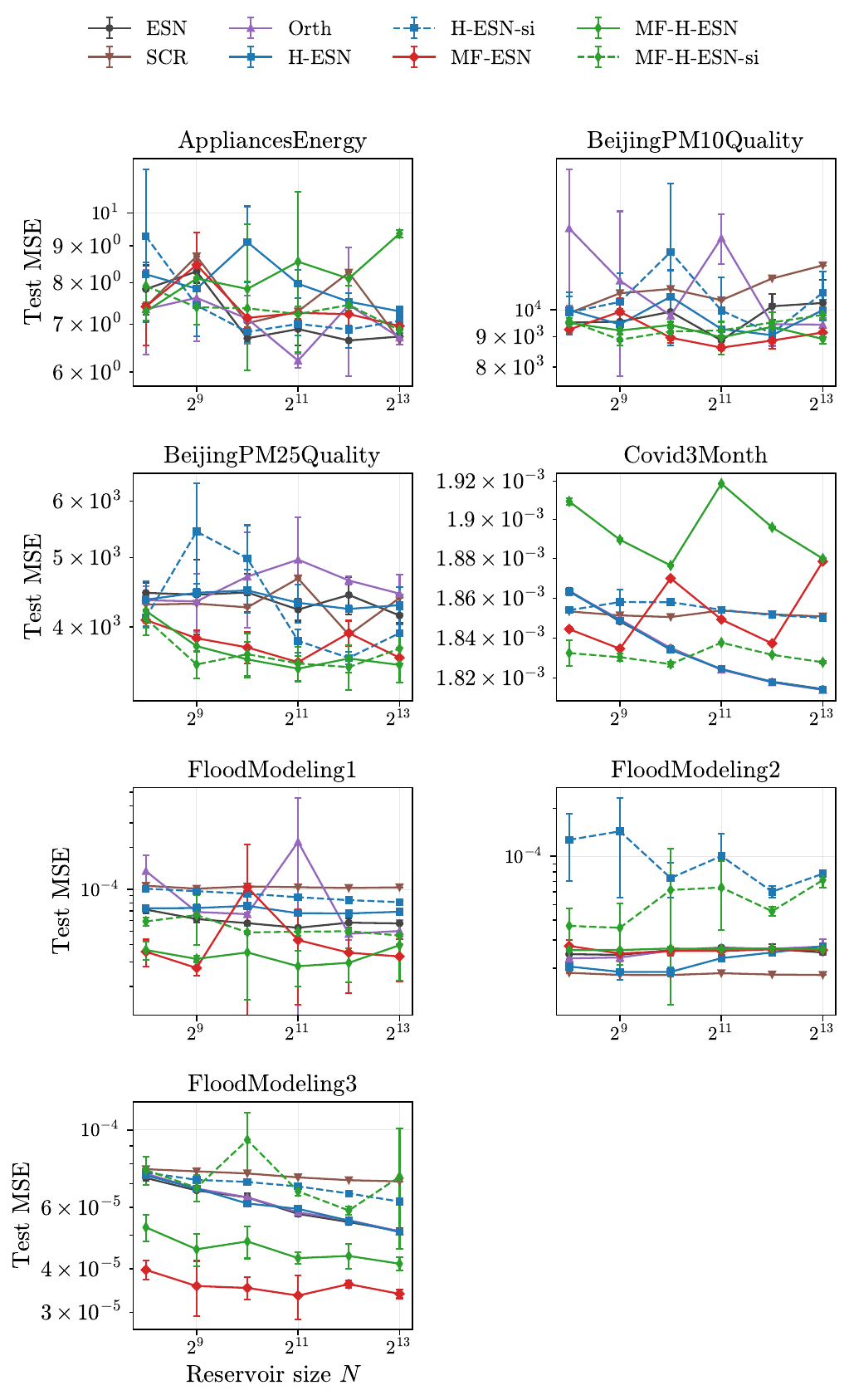}
\caption{Test mean squared error (mean $\pm$ std over 3 instantiations,
logarithmic scale) versus reservoir size, one panel per regression
benchmark. Lower is better. Note the different vertical ranges: on
Covid3Month all models lie within a few percent of each other, as discussed
in Section~\ref{sec:fixedN}.}
\label{fig:scaling_all_reg}
\end{figure}

Figure~\ref{fig:scaling_all_reg} does the same for the seven regression
benchmarks. It makes visible what Section~\ref{sec:scaling} reports in
aggregate. On most of these tasks the error is essentially flat in $N$, and
on Covid3Month the whole vertical range spans less than $6\%$ of the error,
so the apparent separations there are not meaningful.

\bibliographystyle{elsarticle-num}
\bibliography{biblio}

\end{document}